\documentclass[journal]{IEEEtran}
\usepackage{amssymb}
\usepackage[figuresright]{rotating}
\usepackage[tableposition=top]{caption}
\usepackage{amsmath,subfigure,epsfig,bbding}
\usepackage{epstopdf}
\usepackage{multicol}
\usepackage{booktabs}
\usepackage{multirow}
\usepackage{threeparttable}
\usepackage{color}
\DeclareCaptionFont{blue}{\color{blue}}
\usepackage{makecell}
\usepackage[linesnumbered,ruled,vlined]{algorithm2e}
\usepackage{algpseudocode}
\usepackage{amsmath}
\usepackage{url}

\begin{document}

\title{StarVQA+: Co-training Space-Time Attention for Video Quality Assessment}

\author{Fengchuang~Xing, Yuan-Gen~Wang,~\IEEEmembership{Senior Member,~IEEE,}
 Weixuan Tang,~\IEEEmembership{Member,~IEEE,}\\
 Guopu~Zhu,~\IEEEmembership{Senior Member,~IEEE}, and Sam~Kwong,~\IEEEmembership{Fellow,~IEEE}

\thanks{

F. Xing, Y.-G. Wang are with the School of Computer Science and Cyber Engineering, Guangzhou University, Guangzhou 510006, China (e-mail: xfchuang@e.gzhu.edu.cn; wangyg@gzhu.edu.cn).

W. Tang is the Institute of Artificial Intelligence and Blockchain, Guangzhou University, Guangzhou 510006, China (E-mail: tweix@gzhu.edu.cn).

G. Zhu is with the School of Computer Science and Technology, Harbin Institute of Technology, Harbin, 150001, China (e-mail: guopu.zhu@hit.edu.cn).

S. Kwong is with the Department of Computer Science, City University of Hong Kong, Hong Kong (e-mail: cssamk@cityu.edu.hk).

}}

\maketitle

\begin{abstract}
Self-attention based Transformer has achieved great success in many computer vision tasks. However, its application to video quality assessment (VQA) has not been satisfactory so far. Evaluating the quality of in-the-wild videos is challenging due to the unknown of pristine reference and shooting distortion. This paper presents a co-trained Space-Time Attention network foR the VQA problem, termed StarVQA+. Specifically, we first build StarVQA+ by alternately concatenating the divided space-time attention. Then, to facilitate the training of StarVQA+, we design a vectorized regression loss by encoding the mean opinion score (MOS) to the probability vector and embedding a special token as the learnable variable of MOS, leading to better fitting of human's rating process. Finally, to solve the data hungry problem with Transformer, we propose to co-train the spatial and temporal attention weights using both images and videos. Various experiments are conducted on the de-facto in-the-wild video datasets, including LIVE-Qualcomm, LIVE-VQC, KoNViD-1k, YouTobe-UGC, LSVQ, LSVQ-1080p, and DVL2021. Experimental results demonstrate the superiority of the proposed StarVQA+ over the state-of-the-art. The source code is publicly available at https://github.com/GZHU-DVL/StarVQAplus.
\end{abstract}


\section{Introduction}

\IEEEPARstart{U}{ser}-generated content (UGC) has grown rapidly over the past few years on popular social media sites, including TikTok, Facebook, Instagram, YouTube, and Twitter \cite{Omnicore2021}. The storage, streaming, and utilization of these UGC have become challenging issues. Currently, the Internet is mostly inundated by a stream of poor-quality videos shot by some amateur videographers.
Therefore, screening these videos via video quality assessment (VQA) is of great necessity. A variety of tasks for assessing video quality have emerged, such as 3D VQA \cite{zhang2019sparse}, animation VQA \cite{xian2022content}, compressed VQA \cite{zhang2021deep,mao2021high}, etc.
However, it is quite difficult to assess the perceptual quality of videos shot in the wild, since neither pristine reference nor shooting distortion is accessible.

The performance of deep convolutional neural network (CNN) on the VQA tasks has been astounding. For instance, Kim \textit{et al.} \cite{kim2018deep} proposed
an aggregation network based on deep CNN (called DeepVQA) to learn the maps of spatiotemporal visual sensitivity.
Zhang \textit{et al.} \cite{Zhang2019} utilized transfer learning to construct a general-purpose VQA system. You and Korhonen \cite{you2019deep} employed a 3D convolution network to extract regional spatiotemporal characteristics from short video clips. Li \textit{et al.} \cite{VSFA, MDVSFA} extracted the spatial characteristics from each frame using the pre-trained ResNet-50. To combine motion information from various temporal frequencies, Chen \textit{et al.} \cite{chen2020rirnet} created a motion perception network.
Ying \textit{et al.} \cite{LSVQ} proposed to extract the 2D and 3D features from two branches, and further combine these features via a time series regressor for video quality prediction.
For leveraging the advantages of both quality-aware scene statistics features and semantics-aware deep convolutional features, Tu \textit{et al.} \cite{tu2021rapique} combined spatiotemporal scene statistics with high-level semantic data.
Unfortunately, because reference videos and specific distortion types are unavailable and the receptive field of convolutional kernels in CNN is constrained \cite{arnab2021vivit}, the aforementioned methods encounter the performance bottleneck on the in-the-wild video datasets \cite{LSVQ, LIVE-VQC, hosu2017konstanz}.

\begin{figure*}[tp]
\centering
\includegraphics[scale=0.60]{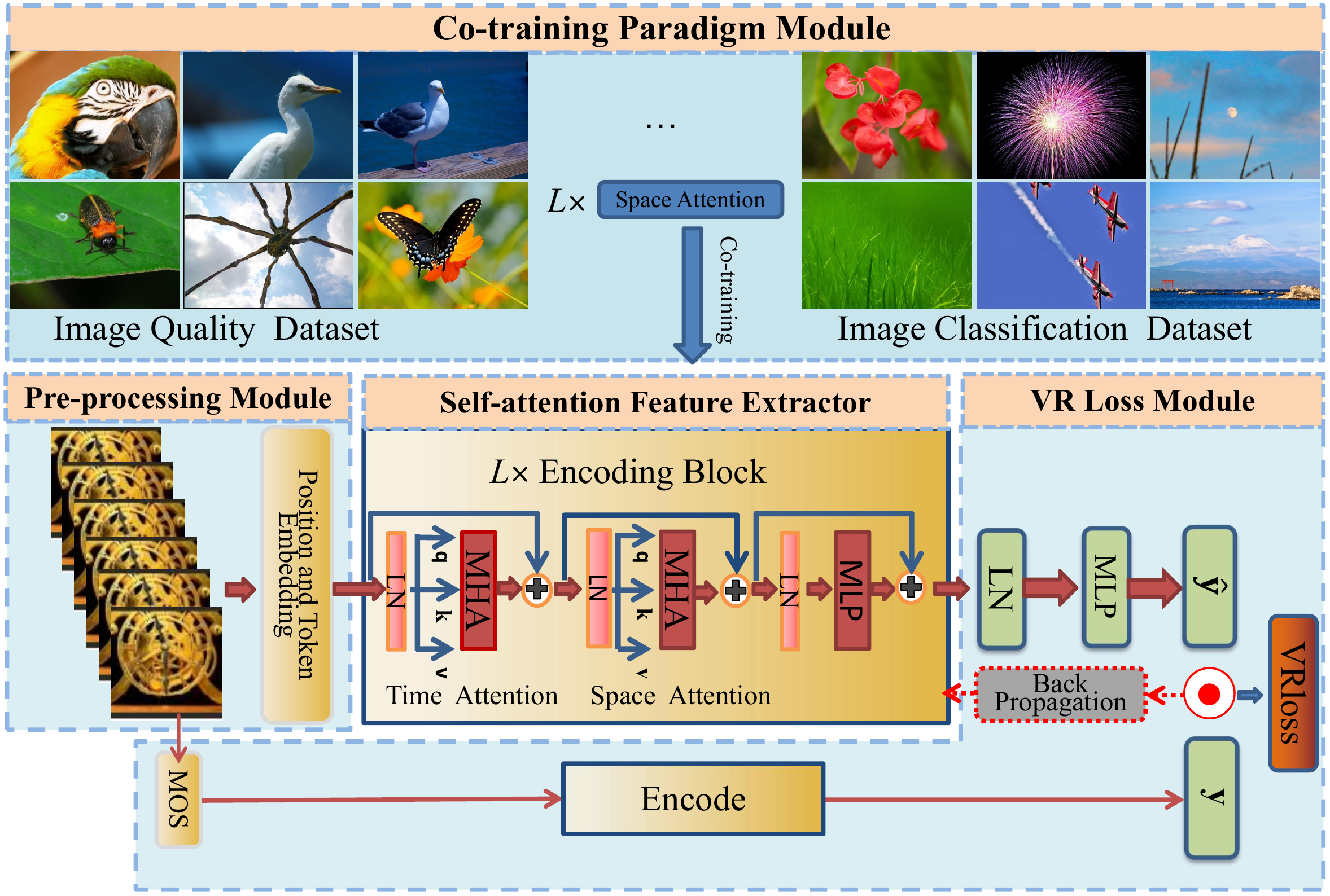}

\caption{Description of StarVQA+ network. It includes pre-processing, encoding block, and vectorized regression loss modules.}\label{fig1}

\end{figure*}

Recent techniques in computer vision have been motivated by the success of the self-attention mechanism in natural language processing (NLP),
wherein Transformers are introduced to either entirely replace CNNs \cite{ramachandran2019stand} \cite{parmar2018image} or be integrated into CNNs \cite{liu2018end} \cite{wang2018non}.
For instance, in image classification tasks,
a pure Transformer-based architecture, called
Vision Transformer (ViT) \cite{dosovitskiy2020image}, has beaten its convolutional competitors, where Transformer does not use any convolutions but is based on multi-headed self-attention \cite{Transformer}. Self-attention mechanism is particularly effective in modeling the long-term dependency of sequential language. Like sentences, videos follow one another. So, one expects that such self-attention will also be useful for video modeling \cite{timesformer}. Inspired by ViT, a number of Transformer-based models were created for video classification tasks \cite{arnab2021vivit,timesformer,videoswin,vidtr}. Compared to convolutional networks, these models achieved a higher classification accuracy. Unlike classification tasks, regression tasks must produce a continuous real value that is as close to the ground truth as possible, which attempt to discriminate between a variety of continuous values. Studies revealed that applying existing classification networks to regression tasks did not work well \cite{dosovitskiy2020image,timesformer,PQA}. As noted by A. C. Bovik \cite{Bovik2020}, ``Unlike human participation in crowdsourced picture labeling experiments like ImageNet, where each human label might need only 0.5-1.0 seconds to apply, human quality judgments on pictures generally required 10-20x that amount to time for a subject to feel comfortable in making their assessments on a Likert scale \cite{Ghadiyaram2016}.'' In general, a video clip has a long duration including hundreds of image frames, and the difference in perceptual quality between it and other videos with different content is extremely subtle. Can the Transformer be applied to VQA task? If yes, how to implement it effectively? These questions become the original motivation of this work.

With the deep investigation on VQA task, various VQA datasets have been developed accordingly, such as CVD2014 \cite{CVD2014}, LIVE-Qualcomm \cite{LIVE-Q}, LIVQ-VQC \cite{LIVE-VQC}, KoNViD-1k \cite{hosu2017konstanz}, LSVQ \cite{LSVQ}, YouTobe-UGC \cite{UGC-2019}, and DVL2021 \cite{DVL2021}. These datasets focus on different aspects, including the difference in resolution and capturing devices, diverse contents and scenes, and ultra-high definition (UHD) video. Simultaneously, with the increase of data and the diversity of video content, the difficulty of creating VQA model has likewise increased. It is known that compared with CNNs, Transformer models are more data-hungry \cite{touvron2021training} and current VQA datasets are relatively small in scale. It is a big challenge to train such large Transformer network using small-scale VQA datasets. Inspired by previous works in vision and language \cite{lu2019vilbert,su2019vl}  that demonstrate a single Transformer model can be extended to many downstream tasks, we propose to leverage both the image and video data to co-train a single Transformer-based VQA model in this paper. We aim to design a co-training strategy for a general purpose VQA model. Such a co-training strategy is buttressed by the following two facts. First, each separate video dataset covers itself rich video content, and thus training a single network on multiple datasets together can make the trained model have good generalization ability. Second, images are one natural source for training space-attention module, while videos are the other for training both the time-attention and space-attention modules. Besides, compared to image datasets, current VQA datasets are relatively smaller in scale. Training Transformer by combining images data with many image frames of the video sequence and leveraging a diverse distribution of image content can be beneficial in building robust spatial representations in VQA models. In summary, the contributions of this work are as follows:

\begin{itemize}
\item[$\bullet$] A novel Transformer network is built, named StarVQA+ in this paper. To our best knowledge, this is the first work to apply pure-Transformer to the VQA problem. This work broadens Transformer to a new application and demonstrates that the self-attention has also excellent potential in the VQA field.

\item[$\bullet$] A vectorized regression loss function (VRloss) is designed, easing the training of StarVQA+. Our designed VRloss enables the existing Transformer models, which are originally designed for classification tasks, to solve the regression problems.

\item[$\bullet$] A new co-training strategy is proposed to solve the data-hungry problem with Transformer. To this aim, we employ both the large-scale image dataset (ImageNet) and video data to co-train the proposed Transformer architecture, alleviating the tension between the existing small-scale VQA datasets and the training of huge Transformer.

\item[$\bullet$] Extensive experiments are performed on the benchmark in-the-wild video datasets and show that the proposed StarVQA+ achieves competitive performance compared with ten state-of-the-art VQA methods.

\end{itemize}

Compared with the preliminary conference version \cite{xing2021starvqa}, the major improvements of this paper come from the following three aspects: 1) We design a new co-training strategy in Section III-B for solving the data-hungry problem with Transformer. 2) We provide the experimental results on more in-the-wild VQA datasets. The results are shown in Tables \ref{train Paradigm} and \ref{compare-table}. Moreover, we present the detailed analysis on the results in Section IV-B. 3) We perform the hyperparameter study and the computational complexity analysis in Section IV-D and in Section IV-E, respectively. In the hyperparameter study, we separately illustrate the performance impact by the number of frames sampled from each video and the number of anchors in VRloss.

The rest of the paper is organized as follows. In Section II, we briefly reviews previous literature relating to VQA models. Section III unfolds the details of the proposed StarVQA+. Experimental results are provided in Section IV, followed by our conclusions in Section V.

\section{Related Work}
In this section, we introduce the related work including classical and deep learning based VQA methods.

\subsection{Classical VQA}
Most early VQA models were distortion-specific, such as blockiness \cite{wang2000blind}, blur \cite{marziliano2002no}, ringing \cite{feng2006measurement}, banding \cite{tu2020bband}, and noise \cite{norkin2018film}.
However, these distortion-specific models are incapable of handling complicated distortions within videos in real scenarios. Afterwards, based on classical machine learning, some high-performing VQA methods were designed.
Within them, the most popular VQA \cite{ruderman1994statistics} used perceptually relevant and low-level features based on natural scene statistics (NSS) models, which were often predictably altered by the presence of distortions \cite{sheikh2006image}.
The NSS models were also explored in the wavelet-domain BIQI \cite{moorthy2010two} and the spatial-domain NIQE \cite{mittal2012making}, and were further extended to encompass natural bandpass space-time video statistics models \cite{li2016spatiotemporal,mittal2015completely,sinno2019spatio,saad2014blind}. Other extensions of the empirical NSS included the joint statistics of the gradient magnitude and Laplacian of Gaussian (GM-LOG \cite{xue2014blind}) in log-derivative and log-Gabor spaces (DESIQUE \cite{zhang2013no}).

The perception of video is correlated with both spatial and temporal changes.
In \cite{li2015no}, Li \textit{et al}. extracted NSS using 3D shearlet transform to make it more discriminative. Regarding the joint modeling of spatiotemporal statistics, the authors in \cite{li2016spatiotemporal} proposed to adopt 3D-DCT transforms of local space-time regions from videos to extract quality-aware features. More recently, Dendi and Channappayya \cite{dendi2020no} conducted 3D divisive normalization transform (DNT) and spatiotemporal Gabor-filtered responses of 3D-DNT coefficients of natural videos. However, the 3D transform is too time-consumed for practical use. And neither of the above models has been observed to perform well on UGC datasets \cite{LIVE-VQC,hosu2017konstanz}. Designing separable spatial-temporal statistics is another interesting and more practical approach to incorporating temporal features into VQA models \cite{sinno2019spatio,yu2021predicting,madhusudana2021st,chen2020perceptual}. Spatial features can be modified to capture temporal effects, like BRISQUE \cite{BRISQUE}, wherein simple frame-differences or spatially displaced frame-differences are deployed \cite{sinno2019spatio,yu2021predicting,lee2020video,lee2021space}.

To enhance the computational efficiency, a very recent feature-based VQA model called TLVQM \cite{TLVQM} used a two-level feature extraction mechanism to achieve efficient computation of a set of impairment/distortion-relevant features. On the natural VQA datasets, VIDEVAL \cite{VIDEVAL} currently achieved very good performance at a reasonable complexity. RAPIQUE \cite{tu2021rapique} designed an effective and efficient video quality model for UGC content. It combines the benefits of both quality-aware scene statistics and semantics-aware features for video quality modeling, yielding top performance at a low computational cost.

\subsection{Deep Learning Based VQA}
With the help of a large number of labeled data, deep convolutional neural networks (CNNs) have been shown to deliver standout performance in a wide variety of computer vision tasks \cite{ying2020patches,chen2020proxiqa,chen2020learning}. Unfortunately, existing VQA datasets, such as CVD2014 \cite{CVD2014}, LIVE-Qualcomm \cite{LIVE-Q}, LIVE-VQC \cite{LIVE-VQC}, KoNViD-1k \cite{hosu2017konstanz}, and YouTube-UGC \cite{UGC-2019}, are still in small-scale and are not enough to train the large networks. To conquer the limits of small-scale dataset, researchers have proposed to either conduct patch-wise data-augmentation during training \cite{kang2014convolutional,bosse2016deep,kim2017deep} or pretrain deep nets on larger visual sets like ImageNet \cite{imagenet} for quality prediction \cite{you2019deep}, \cite{VSFA}, \cite{MDVSFA}, \cite{liu2018end}, which report remarkable performance on naturally distorted datasets \cite{LSVQ}, \cite{LIVE-VQC}, \cite{UGC-2019}, \cite{hosu2020koniq}.

Inspired by the idea of pretraining, many deep CNN models have developed to capture the spatiotemporal information of the video by pretraining on ImageNet. For example, Kim \textit{et al}. \cite{kim2018deep} proposed a deep video quality assessor (DeepVQA) to learn spatio-temporal visual sensitivity maps via a deep CNN and a convolutional aggregation network. V-MEON \cite{liu2018end} built a multi-task CNN framework by jointly optimizing a 3D-CNN for feature extraction and a codec classifier. You and Korhonen \cite{you2019deep} designed a VQA model with a 3D-CNN as the feature extractor and a Long-Short Term Memory (LSTM) for the overall quality prediction. VSFA \cite{VSFA} applied a pre-trained image classification CNN as a deep feature extractor, and then integrated the frame-wise deep features using a gated recurrent unit, reporting excellent performance on several natural video datasets \cite{hosu2017konstanz,CVD2014,ghadiyaram2017capture}. They further built an enhanced version of VSFA, dubbed MDTVSFA \cite{MDVSFA}, by performing  the training strategy on mixed datasets to train a single VQA model. Hou \textit{et al}. \cite{hou2020no} used VGG-net pretrained on ImageNet to compute features of video frames and then the overall video quality was predicted by a 3D-CNN model based on the features. Similar pretraining strategies have been employed in \cite{varga2019no}, \cite{goring2018deviq}, \cite{VSFA}, \cite{varga2019no2}, \cite{korhonen2020blind}, where different CNN architectures with ImageNet pretrained weights (ResNet-50 in \cite{VSFA} and \cite{korhonen2020blind}, InceptionV3 in \cite{goring2018deviq}) were employed to compute image features on video frames. These features were fed into RNNs (e.g., LSTM in \cite{varga2019no} and \cite{korhonen2020blind}, gated recurrent units in \cite{VSFA}) or other regressors (e.g., random forest in \cite{goring2018deviq}) to obtain the prediction of video quality. The above VQA models all used ImageNet to pretrain CNNs, even though ImageNet is not an intention of VQA dataset. Ying \textit{et al}. \cite{LSVQ} created a local-to-global region-based VQA architecture using a DNN that computes both 2D and 3D video features. Li \textit{et al}. \cite{li2022blindly} proposed to transfer knowledge from image quality assessment (IQA) datasets with authentic distortions and the large-scale action recognition dataset with rich motion patterns. A promising architecture LSCT-PHIQNet \cite{you2021long} was proposed by modeling the perceptual mechanisms on IQA and combining CNN and Transformer for the VQA problem.

\section{Proposed Method}
In this paper, matrices, vectors, and scalar variables are in bold uppercase, bold lowercase, and italic lowercase, respectively. $|C|$, $\lfloor\cdot\rfloor$, $(\cdot)^T$, $\mathbf{LN}(\cdot)$, $\mathbf{SM}(\cdot)$, $\langle\cdot\rangle$, $[\mathbf{a},\mathbf{B}]$, and $||\cdot||$ denote the integer set $\{1, \dots, C\}$, the floor operation, the transpose operation, the LayerNorm function, the Softmax activation function, the inner product operation, the operation of concatenating vector $\mathbf{a}$ and matrix $\mathbf{B}$ into a matrix, and $L_2$-norm, respectively. The overall framework of the proposed StarVQA+ is shown in Fig. \ref{fig1}, which consists of two key components: a Transformer-based space-time attention network and a co-training paradigm.

\subsection{Network Architecture}

In this paper, the architecture of pure-Transformer is introduced into VQA for the first time, and a Transformer-based VQA model, called StarVQA+, is proposed. Our StarVQA+ consists of four modules, including a preprocessing module, a feature extractor made up of time-attention module and space-attention module, and a vectorized regression loss module.
Firstly, the preprocessing module is responsible for converting the divided non-overlapping patches within video frames as initialized embedding tokens.
Afterwards, the feature extractor processes the embedding tokens with multiple successive blocks, wherein each block applies the time-attention module to capture the temporal information between video frames, and applies the space-attention module to capture the spatial information within video frames.
Finally, to adapt the Transformer, which is originally designed for classification tasks, to regression task in VQA, the vectorized regression loss module is designed to construct the loss function.

\subsubsection{Pre-processing Module}

Preprocessing module is responsible for efficiently and comprehensively constructing the abstract representation for a given video, which can be fed into the Transformer architecture.
Note that Transformer is originally applied to NLP, wherein each word is represented by a token.
In our StarVQA+, the divided non-overlapping patches within video frames are converted into tokens, via which high-level spatiotemporal information can be extracted.

In order to reduce the redundant information of different frames within a video, $N$ frames are selected from the video sequence according to equal-interval sampling. Note that $N=1$ if the input is a still image.
Then, each selected frame is cropped to size $H\times W\times 3$ in a random way, where $H$ and $W$ denote the height and width of the cropped frame respectively, and 3 refers to the number of color channels.
By means of training for dozens epoches, the randomly cropped images should cover most semantic information of the original frames.
Compared with image resizing which aims at preserving the semantic information, image cropping can better maintain the image quality, which is suitable for VQA.
The effectiveness of image cropping is also supported by previous works, such as VSFA \cite{VSFA}, 3D-CNN+LSTM \cite{you2019deep}, MDTVSFA \cite{MDVSFA}, RAPIQUE \cite{tu2021rapique}, PVQ \cite{LSVQ}, and BVQA-2021 \cite{li2022blindly}.

Subsequently, the cropped video frame is divided into non-overlapping $ P=\lfloor H/S \rfloor \times \lfloor W/S\rfloor$ patches with size $S \times S$.
Each patch is further flattened into a token with length $D=S\times S\times 3$.
Specifically, $\mathbf{x}_{(p, n)}\in \mathbb{R}^{D}$ denotes the initial token for the $p$-th patch in the $n$-th selected frame, where $p\in|P|$ and $n\in|N|$.
Therefore, such video has $ K= P \times N$ tokens in total.

Then, for the sake of capturing the long-range dependence of spatiotemporal information via self-attention mechanism, the spatiotemporal position information is further embedded into the obtained token as
\begin{equation}
 \mathbf{e}_{(p,n)}^{(0)}=\mathbf{M}\mathbf{x}_{(p,n)}+\mathbf{p}_{(p,n)},
\end{equation}
where $\mathbf{p}_{(p,n)}\in \mathbb{R}^D$ denotes the spatiotemporal position vector, $\mathbf{M}\in \mathbb{R}^{D\times D}$ denotes the learnable matrix, and $\mathbf{e}_{(p,n)}^{(0)}\in \mathbb{R}^{D}$ denotes the embedding token.

Finally, all $\mathbf{e}_{(p,n)}^{(0)}$ from the same video are shaped into a matrix, which is further concatenated with the first position of a special initialized MOS token $\mathbf{e}_{(0,0)}^{(0)}$ as
\begin{equation}\label{V2M}
\begin{split}
\mathbf{E}^{(0)}=\left[
 \begin{array}{c}
 \mathbf{e}_{(0,0)}^{(0)},
 \mathbf{e}_{(1,1)}^{(0)},
  \dots,
   \mathbf{e}_{(P,N)}^{(0)}
   \end{array}
   \right] ,
\end{split}
\end{equation}
where $\mathbf{E}^{(0)}  \in \mathbb{R}^{D \times (K+1)}$.
Note that the elements in $\mathbf{E}^{(0)}$ are denoted as $\mathbf{e}_{(\tilde{p},\tilde{n})}^{(0)}$, where $\tilde{p} \in \{0, \dots,P\}$, $\tilde{n} \in \{0, \dots,N\}$, $(\tilde{p},\tilde{n})\neq (0,n)$, and $(\tilde{p},\tilde{n})\neq (p,0)$.
The superscript 0 in $\mathbf{e}_{(\tilde{p},\tilde{n})}^{(0)}$ indicates that such elements are the input of the first block in the time attention module in the feature extractor. Without loss of generality, the input of the $i$-th block in the time-attention module is denoted as $\mathbf{e}_{(\tilde{p},\tilde{n})}^{(i-1)}$, where $i>0$.

\subsubsection{Self-attention Feature Extractor}
Video is composed of spatial and temporal information.
In our proposed StarVQA+, a multi-head self-attention feature extractor \cite{timesformer} is applied to extract the spatialtemporal information of video.
Such feature extractor consists of multiple encoding blocks, wherein each block applies the time-attention module and space-attention module in a sequential manner.

\paragraph{Query-Key-Value Computation}
Both the time-attention module and space-attention module in the feature extractor applies self-attention based on query, key, and value.
Their calculation process is given as follows.
Note that the feature extractor contains $I$ blocks, and each block applies multi-head attention with $A$ heads.
The given video has $N$ frames, and each frame is divided into $P$ patches.
Specifically, the $a$-th head ($a\in|A|$) in the $i$-th ($i\in|I|$) encoding block is utilized to extract the embedding for the input token $\mathbf{z}$ in the $\tilde{n}$-th frame and the $\tilde{p}$-th patch as
\begin{equation}
\label{eq:q}
 \mathbf{q}_{(\tilde{p},\tilde{n})}^{(i,a)}=\mathbf{W}_{Q}^{(i,a)}\textrm{LN}(\mathbf{z}),
\end{equation}
\begin{equation}
\label{eq:k}
 \mathbf{k}_{(\tilde{p},\tilde{n})}^{(i,a)}=\mathbf{W}_{K}^{(i,a)}\textrm{LN}(\mathbf{z}),
\end{equation}
\begin{equation}
\label{eq:v}
 \mathbf{v}_{(\tilde{p},\tilde{n})}^{(i,a)}=\mathbf{W}_{V}^{(i,a)}\textrm{LN}(\mathbf{z}),
\end{equation}
where $\mathbf{q}_{(\tilde{p},\tilde{n})}^{(i,a)} \in \mathbb{R}^{H}$,
$\mathbf{k}_{(\tilde{p},\tilde{n})}^{(i,a)} \in \mathbb{R}^{H}$, and
$\mathbf{v}_{(\tilde{p},\tilde{n})}^{(i,a)} \in \mathbb{R}^{H}$ denotes the query, key and value, respectively, $\mathbf{W}_{Q}^{(i,a)}\in \mathbb{R}^{H\times D}$, $\mathbf{W}_{K}^{(i,a)} \in \mathbb{R}^{H \times D}$, $\mathbf{W}_{V}^{(i,a)}\in \mathbb{R}^{H \times D}$ denotes the learnable query, key, and value matrices, respectively,
$\mathbf{z}$ denotes the embedding token, and $H=D/A$ denotes the latent dimension in attention head.

\paragraph{Time-attention Module}

Firstly, the time-attention module calculates the query, key, and value according to Eqs. \eqref{eq:q}, \eqref{eq:k}, and \eqref{eq:v}, wherein $\mathbf{z}=\mathbf{e}_{(\tilde{p},\tilde{n})}^{(i-1)}$ in these equations.

Then, the time-attention module calculates the time-attention weights $\boldsymbol{\alpha}_{(\tilde{p},\tilde{n})}^{(i,a)(time)}$ for token $\mathbf{e}_{(\tilde{p},\tilde{n})}^{(i-1)}$ by means of comparing the query of $\mathbf{e}_{(\tilde{p},\tilde{n})}^{(i-1)}$ and the keys of different tokens, including the MOS token and
those tokens in the same position but different frames, as
\begin{equation}\label{tweight}
 \boldsymbol{\alpha}_{(\tilde{p},\tilde{n})}^{(i,a)(\textrm{time})}=\mathbf{SM}\left(\frac{\left(\mathbf{q}_{(\tilde{p},\tilde{n})}^{(i,a)(\textrm{time})}\right)^T}{\sqrt{H}}\cdot\left[\mathbf{k}_{(0,0)}^{(i,a)(\textrm{time})}, \mathbf{K}_{(\tilde{p})}^{(i,a)\textrm{(time)}} \right]\right),
\end{equation}
where $\boldsymbol{\alpha}_{(\tilde{p},\tilde{n})}^{(i,a)(\textrm{time})} \in \mathbb{R}^{N+1}$ denotes the time-attention weights,
$\mathbf{K}_{(\tilde{p})}^{(i,a)(\textrm{time})} \in \mathbb{R}^{H \times  N}$ denotes a matrix formed by $\mathbf{k}_{(p,n)}^{(i,a)(\textrm{time})}$
with $p=\tilde{p}$ and different $n$ as
\begin{equation}\label{V2M}
\begin{split}
\mathbf{K}_{(\tilde{p})}^{(i,a)\textrm{(time)}}=\left[
 \begin{array}{c}
 \mathbf{k}_{(\tilde{p},1)}^{(i,a)(\textrm{time})},
  \dots,
   \mathbf{k}_{(\tilde{p}, N)}^{(i,a)(\textrm{time})}
   \end{array}
   \right].
\end{split}
\end{equation}

Afterwards, the values of different tokens are weighted by their corresponding time-attention weights as
\begin{equation}
\begin{split}
 \mathbf{s}_{(\tilde{p},\tilde{n})}^{(i,a)(\textrm{time})}=\mathbf{v}_{(0,0)}^{(i,a)(\textrm{time})} \boldsymbol{\alpha}_{(0,0)}^{(i,a)(\textrm{time})}(0)+ \\ \sum_{\tilde{n}'=1}^{ N}\mathbf{v}_{(\tilde{p},\tilde{n}')}^{(i,a)(\textrm{time})} \boldsymbol{\alpha}_{(\tilde{p},\tilde{n}))}^{(i,a)(\textrm{time})}(\tilde{n}'),
 \end{split}
\end{equation}
where both $\boldsymbol{\alpha}_{(0,0)}^{(i,a)(\textrm{time})}(0)$ and $\boldsymbol{\alpha}_{(\tilde{p},\tilde{n}))}^{(i,a)(\textrm{time})}(\tilde{n}')$ denotes scalars within $\boldsymbol{\alpha}$, which are utilized to multiply each elements within vector $\mathbf{v}$, and $\mathbf{s}_{(\tilde{p},\tilde{n})}^{(i,a)(\textrm{time})} \in \mathbb{R}^{H}$ denotes the encoding coefficients of the time-attention module.

The encoding coefficients from different heads are concatenated, and are further projected as
\begin{equation}
\begin{split}
 \mathbf{\tilde{e}}_{(\tilde{p},\tilde{n})}^{(i)}=\mathbf{W}_{T}\left[
 \begin{array}{c}
 \mathbf{s}_{(\tilde{p},\tilde{n})}^{(i,1)(\textrm{time})}\\
  \vdots \\
   \mathbf{s}_{(\tilde{p},\tilde{n})}^{(i,A)(\textrm{time})}
   \end{array}
   \right]
   +\mathbf{e}_{(\tilde{p},\tilde{n})}^{(i-1)},
\end{split}
\end{equation}
where $\mathbf{\tilde{e}}_{(\tilde{p},\tilde{n})}^{(i)} \in \mathbb{R}^{D}$ denotes the output of the time-attention module, $\mathbf{W}_{T}\in \mathbb{R}^{D\times D}$ denotes the learnable mapping matrix.

\paragraph{Space-attention Module}

Based on the output of the time-attention module, the space-attention module calculates query, key, and value according to Eqs. \eqref{eq:q}, \eqref{eq:k}, and \eqref{eq:v}, wherein $\mathbf{z}=\mathbf{\tilde{e}}_{(\tilde{p},\tilde{n})}^{(i)}$ in these equations.

Then, the space-attention module calculates the space-attention weights $\boldsymbol{\alpha}_{(\tilde{p},\tilde{n})}^{(i,a)(spa)}$ for token $\mathbf{\tilde{e}}_{(\tilde{p},\tilde{n})}^{(i)}$ by means of comparing the query of $\mathbf{\tilde{e}}_{(\tilde{p},\tilde{n})}^{(i)}$ and the keys of different tokens, including the MOS token and those tokens in the same frame but different positions, as

\begin{equation}\label{sweight}
 \boldsymbol{\alpha}_{(\tilde{p},\tilde{n})}^{(i,a)(\textrm{spa})}=\mathbf{SM}\left(\frac{\left(\mathbf{q}_{(\tilde{p},\tilde{n})}^{(i,a)(\textrm{spa})}\right)^T}{\sqrt{H}}\cdot\left[\mathbf{k}_{(0,0)}^{(i,a)(\textrm{spa})}, \mathbf{K}_{(\tilde{n})}^{(i,a)(\textrm{spa})} \right]\right),
\end{equation}
where $ \boldsymbol{\alpha}_{(\tilde{p},\tilde{n})}^{(i,a)(\textrm{spa})}  \in \mathbb{R}^{P+1} $ denotes the space-attention weights, $\mathbf{K}_{(\tilde{n})}^{(i,a)(\textrm{spa})} \in \mathbb{R}^{H \times P} $ denotes a matrix formed by $\mathbf{k}_{(p,n)}^{(i,a)(\textrm{spa})}$ with $n=\tilde{n}$ and different $p$ as
\begin{equation}\label{V2M}
\begin{split}
\mathbf{K}_{(\tilde{n})}^{(i,a)\textrm{(spa)}}=\left[
 \begin{array}{c}
 \mathbf{k}_{(1,\tilde{n})}^{(i,a)(\textrm{spa})},
  \dots,
   \mathbf{k}_{( P, \tilde{n})}^{(i,a)(\textrm{spa})}
   \end{array}
   \right] ,
\end{split}
\end{equation}

Next, the values of different tokens are weighted by their corresponding space-attention weights as
\begin{equation}
\begin{split}
 \mathbf{s}_{(\tilde{p},\tilde{n})}^{(i,a)(\textrm{spa})}=\mathbf{v}_{(0,0)}^{(i,a)(\textrm{spa})} \boldsymbol{\alpha}_{(0,0)}^{(i,a)(\textrm{spa})}(0)+  \\  \sum_{\tilde{p}'=1}^{ P}\mathbf{v}_{(\tilde{p}',\tilde{n})}^{(i,a)(\textrm{spa})} \boldsymbol{\alpha}_{(\tilde{p},\tilde{n})}^{(i,a)(\textrm{spa})}(\tilde{p}'),
 \end{split}
\end{equation}
where both $\boldsymbol{\alpha}_{(0,0)}^{(i,a)(\textrm{spa})}(0)$ and $\boldsymbol{\alpha}_{(\tilde{p},\tilde{n}))}^{(i,a)(\textrm{spa})}(\tilde{p}')$ denote scalars within $\boldsymbol{\alpha}$, which are utilized to multiply each elements within vector $\mathbf{v}$, and $\mathbf{s}_{(\tilde{p},\tilde{n})}^{(i,a)(\textrm{spa})} \in \mathbb{R}^{H}$ denotes the encoding coefficients of the space-attention module. The encoding coefficients from different heads are concatenated, and are further projected as
 \begin{equation}
\begin{split}
 \mathbf{\hat{e}}_{(\tilde{p},\tilde{n})}^{(i)}=\mathbf{W}_{S}\left[
 \begin{array}{c}
 \mathbf{s}_{(\tilde{p},\tilde{n})}^{(i,1)(spa)}\\
  \vdots \\
   \mathbf{s}_{(\tilde{p},\tilde{n})}^{(i,A)(spa)}
   \end{array}
   \right]
   +\mathbf{\tilde{e}}_{(\tilde{p},\tilde{n})}^{(i)},
\end{split}
\end{equation}
where $\mathbf{\hat{e}}_{(\tilde{p},\tilde{n})}^{(i)} \in \mathbb{R}^{D}$ denotes the output of space-attention module, $\mathbf{W}_{S} \in \mathbb{R}^{D\times D} $ denotes the learnable mapping matrix.

Finally, $\mathbf{\hat{e}}_{(\tilde{p},\tilde{n})}^{(i)}$ is fed into a Multilayer Perceptron (MLP) as:
\begin{equation}
 \mathbf{\bar{e}}_{(\tilde{p},\tilde{n})}^{(i)}=\mathbf{MLP}\left(\mathbf{LN}\left(\mathbf{\hat{e}}_{(\tilde{p},\tilde{n})}^{(i)}\right)\right)+\mathbf{\hat{e}}_{(\tilde{p},\tilde{n})}^{(i)}.
\end{equation}
where $\mathbf{\bar{e}}_{(\tilde{p},\tilde{n})}^{(i)} \in \mathbb{R}^{D}$ denotes the output of the $i$-th block in the feature extractor.
To this end, the last block of the feature extractor
outputs $\mathbf{\bar{e}}_{(\tilde{p},\tilde{n})}^{(I)} \in \mathbb{R}^{D}$.

\subsubsection{Vectorized Regression Loss Module}

In the early experiments, we attempt to end-to-end train Transformer with existing loss functions, such as $L_2$-norm, hinge, and cross-entropy loss, but such Transformer achieves rather poor VQA performance.
Detailed results can be referred to Section \ref{anchor}. Therefore, we delicately design a vectorized regression loss module, and its key idea is
to transform the quality score into a probability vector.

\begin{table*}[ht]
  \renewcommand\arraystretch{1.2}
   \centering
  \caption{Summary of the datasets used in this paper. Here, ``-'' denotes ``Unavailable''.}\label{dataset}
  \begin{tabular}{p{45pt}cccccccccccc}
   \Xhline{1pt} %
   DATASET &&  YEAR & CONT & TOTAL & RESOLUTION   & FR & LEN  & FORMAT  & MOS Range & SUBJ   &ENV \\
   \hline %
    CVD2014    && 2014   & 5    &  234  & 720p, 480p  & 9-30  & 10-25 & AVI   & [-6.50, 93.38]  & 210         & In-lab \\    
    KoNViD-1k  && 2017   & 1200 &  1200 & 540p       & 24-30 & 8     & MP4   & [1.22, 4.64]  & 642     & Crowd \\    
    LIVE-Qualcomm   && 2018   & 54   &  208 & 1080p  & 30 & 15 & YUV & [16.5621, 73.6428] & 39   & In-lab \\    
    LIVE-VQC   && 2018   & 585   &  585 & 240p-1080p  & 19-30 & 10 & MP4 & [6.2237, 94.2865] & 4776   & Crowd \\    
    YouTube-UGC   && 2019   & 1380   &  1380 & 360p-4k  & 15-60 & 20 & MP4 & [1.242, 4.698] & $>$8K  & Crowd \\    
    LSVQ   && 2020   & 39075   &  39075 & 99p-4k  & 25/50 & 5-12 & MP4 & [2.4483, 91.4194] & 5.5M   & Crowd \\    
    DVL2021   &&  2021   & 206   &  206 & 4k  & 50 & 8 & MP4 & [1.00, 4.84] & 32   & In-lab \\    %
    \hline %
    KonIQ-10k && 2020  & 10073   &10073 &-- &-- &-- &-- & [1.096, 4.310]&1459  &Crowd\\
    SPAQ && 2020  & 11125   &11250 &-- &-- &-- &-- & [2.0, 96.0]&1459  &Crowd\\

   \Xhline{1pt}
  \end{tabular}
   \begin{tablenotes}
 \item[1] CONT: Total number of unique contents. TOTAL: Total number of test sequences including reference and distorted videos.
 \item[2] FR: Frame rate. LEN: Video duration/length (in seconds). FORMAT: Video container.
 \item[3] RESOLUTION: Video resolution. MOS Range: Range of mean opinion score. SUBJ: Total number of subjects in the study.
 \item[4] ENV: Subjective testing environment. In-lab: Study was conducted in a laboratory. Crowd: Study was conducted by crowdsourcing.
 \end{tablenotes}
 \end{table*}

\begin{table*}[htp]
\centering
\caption{The performance comparison among the four training policies.}\label{train Paradigm}
\begin{tabular}{lccccccccccccc}
\Xhline{1pt}%
Training Dataset  &\multicolumn{2}{c}{KoNViD-1k}    &\multicolumn{2}{c}{DVL2021}     &\multicolumn{2}{c}{LIVE-VQC}   &\multicolumn{2}{c}{LIVE-Qualcomm}  &\multicolumn{2}{c}{YouTobe-UGC}   &\multicolumn{2}{c}{LSVQ}\\
\hline %
Pretrain Dataset & SROCC & PLCC       & SROCC & PLCC                  &SROCC & PLCC                    & SROCC & PLCC                         & SROCC & PLCC                 & SROCC & PLCC  \\ \cline{2-13}

   None   &0.4211 &0.4826                &0.5114 &0.5672                     &0.4201 &0.4918                      &0.5476 &0.5254            &0.7001 &0.7016                  &0.6031 &0.6028\\
   IQA    &0.6334 &0.6230                &0.6111 &0.6235                     &0.4815 &0.4742                      &0.6136 &0.5654            &0.7226 &0.7215                  &0.6244 &0.6231\\
   ImageNet-1k  &0.7901 &0.7806          &0.6652 &0.6643                     &0.7221 &0.7343                      &0.6665 &0.6642            &0.7436 &0.7427                  &0.8532 &0.8563\\
   ImageNet-21k &0.7912 &0.7847          &0.6752 &0.6821                     &0.7543 &0.7433                      &0.6754 &0.6828            &0.7621 &0.7733                  &0.8545 &0.8523\\
\Xhline{1pt} %
\end{tabular}
\end{table*}



From the side of the predicted quality score, note that $\mathbf{\bar{e}}_{(0,0)}^{(I)}$ is the final embedded MOS token in the feature extractor, and is taken as input for the vectorized
regression loss module.
Afterwards, $\textrm{MLP}$ and $\textrm{SM}$ are employed to generate a probability vector according to $\mathbf{\bar{e}}_{(0,0)}^{(I)}$ as
\begin{equation}
 \mathbf{\hat{y}}=\mathbf{SM}\left(\mathbf{MLP}\left(\mathbf{\bar{e}}_{(0,0)}^{(I)}\right)\right),
\end{equation}
where $\mathbf{\hat{y}} \in \mathbb{R}^{M}$ denotes the probability vector of the predicted quality
score,
$M$ denotes the number of anchor point.

From the other side of the ground truth quality score,
for those videos within the same dataset, the MOS of a specific video is linearly scaled into interval [$L$, $U$] as $c$, where $L$ and $U$ represents the lowest and highest scaled MOS within the dataset.
Then, the $i$-th anchor can be calculated as
\begin{equation}
b_i=L+\frac{U-L}{M-1}i,
\end{equation}
where $i={0,1,...M-1}$.
Afterwards, the scaled MOS can be encoded into a probability vector $\mathbf{y} \in \mathbb{R}^{M} =[y_0,\dots,y_{M-1}]$ where
\begin{equation}\label{mos}
 {y}_{i}=\frac{{e}^{(-|c-b_{i}|^2)}}{\sum_{k=0}^{M-1}{e}^{(-|c-b_{k}|^2)}}.
\end{equation}
To this end, the predicted quality score is transformed into $\mathbf{\hat{y}}$, and the ground truth quality score is transformed into $\mathbf{y}$.
Finally, the Vectorized Regression (VR) loss is calculated as
\begin{equation}\label{loss}
\mathcal{L}_{\textrm{VR}}=1-\frac{\langle\mathbf{y}\cdot\mathbf{\hat{y}}\rangle}{||\mathbf{y}||\cdot||\mathbf{\hat{y}}||}.
\end{equation}
In the inference stage, a Support Vector Regressor (SVR) is utilized to reversely transform
the predicted probability vector $\mathbf{\hat{y}}$ into the quality score $\hat{c}$.
Such SVR is trained with training samples ($\mathbf{y}$,$c$).

\subsection{Co-training Paradigm}
Transformer network has a huge volume of parameters, and thus its training generally requires a huge amount of labeled data.
However, constructing VQA datasets are rather time-consuming and labor-consuming. And the sample sizes of most existing VQA datasets are relatively small, such as CVD2014 (234 video samples), LIVE-Qualcomm (208 video samples), LIVE-VQC (585 video samples), and so on.
To overcome the limitation of insufficient training data, in the pre-training stage, a co-training paradigm is proposed to co-train the proposed StarVQA+ using both images and videos across different types of datasets.
The foundation of the co-training paradigm is that in StarVQA+, the time-attention module and space-attention module are decoupled, and therefore can be respectively trained with different datasets.
On the one hand, images contain rich spatial information, and thus are beneficial to pre-training the space-attention module.
On the other hand, videos contain both temporal and spatial information, and thus can be utilized to pre-train both space-attention and time-attention modules. Considering the proper size of KoNViD-1k dataset, it is applied for training and testing in the following experiments in this subsection unless otherwise specified.

\subsubsection{Training space-attention module by images}  \label{image}
It is known that there is a certain correlation and similarity between the image and video data, the IQA and VQA tasks. And considering the fact that the IQA dataset has a huge samples compared to that of VQA, in this part, we focus on pre-training the space-attention module of StarVQA+ with image dataset.
We first construct a weakened version of StarVQA+, whose time-attention module is removed.
Such weakened network is pre-trained on different image datasets, and then its pre-processing module and space-attention module are transferred to StarVQA+.

Specifically, three different datasets are utilized for pre-training.
The first dataset is named as the SPAQ and KonIQ (SK) dataset, which is blended from two large IQA datasets, including SPAQ \cite{SPAQ} and KonIQ-10k \cite{koniq-10k}.
In this case, the weakened network can be pre-trained on the SK dataset for regression task of IQA.
The second and third datasets are ImageNet-1k and ImageNet-21k \cite{imagenet}.
In this case, the vectorized regression loss module in the weakened network is modified as the cross-entropy loss, and such network turns into ViT \cite{dosovitskiy2020image}.
Such network is pre-trained on these datasets for classification task.
Experimental results are shown in Table \ref{train Paradigm}, and the following observations can be made.
\begin{itemize}
\item[$\bullet$] In the absence of the pre-trained image data, the performance obtained by our network is rather poor. The reason is that
    a single VQA dataset is not sufficient for training the Transformer network with a huge volume of parameters.
    By contrast, pre-training the space-attention module of the Transformer with image datasets is beneficial to obtaining better performance in VQA task.

\item[$\bullet$]
Although the SK dataset shares the correlation and similarity with those VQA datasets, pre-training on SK is inferior to pre-training on ImageNet. The reason is that the SK dataset with around 20,000 samples is still far from fully training a Transformer network.
On the other hand, although ImageNet is not originally constructed for VQA task,
these images have great differences in categories, texture, pattern, color, and so on,
which can be utilized to learn the spatial information for the space-attention module of StarVQA+.

\item[$\bullet$]
Pre-training on ImageNet-21k does not bring significant improvement over that on ImageNet-1k. Therefore, we decide to pre-train the space-attention module on ImageNet-1k in the following experiments.
\end{itemize}

\begin{table*}[ht]
\centering
\caption{The performance comparison among different spatiotemporal-attention pre-training policies.}\label{time-attention2}
\begin{tabular}{llll}
\Xhline{1pt}%
Testing Dataset &\multicolumn{3}{c}{KoNViD-1k}\\
\hline %
Space-Attention & Time and Space-Attention                          & SROCC    & PLCC  \\
\hline %
\multirow{9}{*}{ImageNet-1k}
  & Kinetics-400                                          &0.7671    &0.7620\\

  & Kinetics-600                                          &0.7740    &0.7723\\

  & Something-Something V2                                &0.7881    &0.7820\\

  & HowTo100M                                             &0.7562    &0.7520\\ \cline{2-4}

  & LSVQ                                                  &0.8645    &0.8722\\

  & LSVQ+LIVE-VQC                                         &0.8654    &0.8705\\

  & LSVQ+LIVE-VQC+DVL2021                                 &0.8671    &0.8722\\

  & LSVQ+LIVE-VQC+DVL2021+LIVE-Qualcomm                   &0.8606    &0.8604\\

  & LSVQ+LIVE-VQC+DVL2021+LIVE-Qualcomm+YouTobe-UGC       &0.8763    &0.8809\\
\Xhline{1pt} %
\end{tabular}
\end{table*}

\begin{table*}[ht]

  \renewcommand\arraystretch{1.2}
   \centering
  \caption{Summary of different combinations of video datasets.}\label{dataset-combination-name}

  \begin{tabular}{ll}
 \Xhline{1pt} %
  Dataset Combination Name &   Dataset          \\
  \hline %
  Dataset Combination-1   & LIVE-VQC+DVL2021+YouTobe-UGC+LIVE-Qualcomm+KoNViD-1k              \\
  Dataset Combination-2   & LSVQ+DVL2021+YouTobe-UGC+LIVE-Qualcomm+KoNViD-1k                      \\
  Dataset Combination-3   & LSVQ+LIVE-VQC+YouTobe-UGC+LIVE-Qualcomm+KoNViD-1k Dataset                       \\
  Dataset Combination-4   & LSVQ+LIVE-VQC+DVL2021+LIVE-Qualcomm+KoNViD-1k                  \\
  Dataset Combination-5   & LSVQ+LIVE-VQC+DVL2021+YouTobe-UGC+KoNViD-1k              \\
  Dataset Combination-6   & LSVQ+LIVE-VQC+DVL2021+YouTobe-UGC+LIVE-Qualcomm                   \\
   \Xhline{1pt} %
  \end{tabular}
 \end{table*}

\begin{table*}[ht]

  \renewcommand\arraystretch{1.2}
   \centering
  \caption{The performance comparison among different pretraining combinations. Here, \underline{X} means that our StarVQA+ only pre-trains the space-attention module by ImageNet-1k.}\label{joint-train}
  \begin{tabular}{lllllll}
   \Xhline{1pt} %
   Space-attention Pretrain &Test Dataset  & SROCC   &PLCC    & Mixed Dataset Pretrain        & SROCC        &PLCC     \\
   \hline %
     \multirow{6}{*}{ImageNet-1k}
        &LSVQ          &  \underline{0.8501}    & \underline{0.8516} &   Dataset Combination-1              &  0.8532 ($\uparrow$ 0.31\% )    & 0.8623 ($\uparrow$ 1.32\% )                \\
        &LIVE-VQC      &  \underline{0.7221}    & \underline{0.7343} &   Dataset Combination-2              &  0.8566 ($\uparrow$ 18.6\% )    & 0.8743 ($\uparrow$ 19.1\% )               \\
        &DVL2021       &  \underline{0.6623}    & \underline{0.6643} &   Dataset Combination-3              &  0.8429 ($\uparrow$ 27.2\% )    & 0.8412 ($\uparrow$ 26.6\% )                \\
        &YouTobe-UGC   &  \underline{0.7442}    & \underline{0.7475} &   Dataset Combination-4              &  0.8258 ($\uparrow$ 11.1\% )    & 0.8199 ($\uparrow$ 9.61\% )            \\
        &LIVE-Qualcomm &  \underline{0.6675}    & \underline{0.6654} &   Dataset Combination-5              &  0.8291 ($\uparrow$ 24.2\% )    & 0.8191 ($\uparrow$ 23.1\% )           \\
        &KoNViD-1k     &  \underline{0.7912}    & \underline{0.7823} &   Dataset Combination-6              &  0.8763 ($\uparrow$ 10.7\% )    & 0.8809 ($\uparrow$ 12.6\% )            \\
   \Xhline{1pt} %
  \end{tabular}
 \end{table*}

\subsubsection{Training space-attention and time-attention module by videos}\label{video}

Since videos possess spatiotemporal duality,
they can
be utilized to pre-train both the time-attention module and space-attention module of the StarVQA+ network.
In the following part of this subsection, we firstly pre-train the space-attention module of StarVQA+ on ImageNet-1k, and then continuously jointly pre-train the space-attention module and time-attention module on two types of video datasets.

The first type is the mixed VQA datasts, which are
made up of KoNViD-1k, DVL2021, LIVE-VQC, LIVE-Qualcomm, YouTobe-UGC, and LSVQ.
In this case, the space-attention module and time-attention module of StarVQA+ are pre-trained on these datasets for regression task of VQA.
The second type is the video classification datasets, including Kinetics-400 \cite{k400}, Kinetics-600 \cite{k600}, Something-Something V2 \cite{ssv2}, and HowTo100M \cite{howto100}. In this case, the vectorized regression loss module in the StarVQA+ network is modified as the cross-entropy loss, and such network turns into TimeSformer \cite{timesformer}. Such network is pre-trained on these datasets for classification task.
Experimental results are shown in Table \ref{time-attention2}, and the following observations can be made.

\begin{itemize}
\item[$\bullet$] The performance of pre-training on the mixed VQA datasets achieves better performance than that on video classification datasets. The reason is that the mixed VQA dataset also contains more than 30,000 video samples, whose scale is comparable to that of the video classification datasets. Meanwhile, the labelled MOS in the mixed VQA datasets can guide the Transformer network to learn more appropriate quality scores.

\item[$\bullet$] As for pre-training on the mixed VQA datasets, the performance gradually improves as the variety of datasets increases.
    The reason is that such strategy can increase the diversity of samples, and is beneficial to training a VQA model in different scenarios.
\end{itemize}

According to the above experiments, it can be observed that the strategy of jointly pre-training the time-attention and space-attention module on VQA datasets can significantly improve the testing performance on KoNViD-1k dataset. In the following, we further verify the effectiveness of such strategy on different datasets.
Among the six datasets, one specific dataset is used for testing, and the other five datasets are used for pre-training.
The settings of datasets are given in Table \ref{dataset-combination-name}, and the experimental results are given in Table\ref{joint-train}.
The following observations can be made.

\begin{itemize}
\item[$\bullet$]
    Taking the space-attention module pre-trained on ImageNet-1k as the baseline, the average performance improvement of jointly pre-training the space-attention module and time-attention module on VQA datasets is 15.5\% and 16.2\% for SROCC and PLCC respectively on six different settings.

\item[$\bullet$]
The performance improvement is negatively correlated with the scale of the testing dataset, i.e.,
more improvement can be obtained on smaller testing datasets, and vice versa.
Therefore, the strategy of jointly pre-training is more significant on smaller testing datasets.

\end{itemize}

From Section \ref{image} and Section \ref{video}, it can be observed that in the pre-training stage, co-training the proposed StarVQA+ with image and video datasets brings significant improvement. The main steps of co-training are as follows.
Firstly, StarVQA+ is weakened into ViT, and its pre-processing module and space-attention module are trained on ImageNet-1k for classification task. Then, both modules are transferred to StarVQA+. Afterwards, the space-attention module and time-attention module in StarVQA+ are jointly trained on the mixed VQA dataset for VQA task.
\begin{table*}[htp]

  \renewcommand\arraystretch{1.2}
   \centering
 \caption{The performance comparison with state-of-the-art. Here, ``-'' denotes ``Unavailable''.}\label{compare-table}
  \begin{tabular}{cccccccccccc}
   \Xhline{1pt} %
   Dataset      &         &\multicolumn{2}{c}{LIVE-VQC}              &\multicolumn{2}{c}{YouTobe-UGC}               &\multicolumn{2}{c}{LIVE-Qualcomm}  &\multicolumn{2}{c}{KoNViD-1k}  &\multicolumn{2}{c}{DVL2021}\\
   \hline %
   Model        &  Pretrain   &SROCC  &PLCC             &SROCC  &PLCC            &SROCC  &PLCC                     &SROCC & PLCC             &SROCC  &PLCC   \\

   \hline %
   VIIDEO       &  No         & 0.0332 &0.2146           & 0.0580  &0.1534         & 0.1267  &-0.0012              & 0.2988  &0.3002          & --       &--              \\
   V-BLIINDS    &  No         & 0.6939 &0.7178           & 0.5590  &0.5551         & 0.5659  &0.5676               & 0.7101  &0.7037          & --       &--              \\
   TLVQM        &  No         & 0.7988 &0.8025           & 0.6693  &0.6590         & 0.7800  &0.8100               & 0.7729  &0.7688          & 0.7512   &0.7896          \\
   VIDEVAL      &  No         & 0.7522 &0.7514           & 0.7787  &0.7733         & --      &--                   & 0.7832  &0.7800          & 0.7554   &0.8365          \\
   \hline 
   VSFA         &  Yes        & 0.6978 &0.7426           & 0.7242  &0.7431         & 0.7083  &0.7474               & 0.7728  &0.7754          & 0.7567   & 0.7725          \\
   3D-CNN+LSTM  &  Yes        & --    &--                  & --     &--            & 0.6870  &0.7920               & 0.8080  &0.8000          & 0.7646    &0.8225         \\
   MDTVSFA      &  Yes        & 0.7382 &0.7728           & --       &--            & 0.8019  &0.8218               & 0.7812  &0.7856          & --       &--           \\
   RAPIQUE      &  Yes        & 0.7548 &0.7863           & 0.7591  &0.7684         & --      &--                   & 0.8031  &0.8175          & 0.6922   & 0.7946           \\
   PVQ          &  Yes        & 0.8270 &0.8370           & --      &--             & --      &--                   & 0.7910  &0.7860          & --       & --    \\
   BVQA-2022    &  Yes        & 0.8412 &0.8415           & \textbf{0.8312}  &0.8194 & 0.8055  &0.8180               & 0.8362  &0.8335         & --       & --       \\
   StarVQA+     &  Yes        & \textbf{0.8566}  &\textbf{0.8743}     & 0.8258 &\textbf{0.8199}    & \textbf{0.8291}  &\textbf{ 0.8191}         & \textbf{0.8763} &\textbf{0.8809}     & \textbf{0.8429} &\textbf{0.8411}             \\
   \Xhline{1pt} %
  \end{tabular}
 \end{table*}

\section{\textbf{Experimental Results and Analysis}} \label{anchor}
The organizations of Section IV are as follows. The experimental settings are firstly described (given in Section IV-A). Then, in order to evaluate the performance of the proposed StarVQA+, various experiments are conducted, including experiments on individual datasets (given in Section IV-B), experiments on cross datasets (given in Section IV-C) , and experiments of scatter plots (given in Section IV-D).
Finally, ablation studies are conducted to verify the effectiveness of different modules (given in Section IV-F).

\subsection{Experimental Setup}

\subsubsection{Datasets description}

The performance of VQA methods is evaluated on three types of datasets.
The first type is the VQA dataset, including CVD2014 \cite{CVD2014}, LIVE-Qualcomm \cite{LIVE-Q}, LIVE-VQC \cite{LIVE-VQC}, KoNViD-1k \cite{hosu2017konstanz}, LSVQ \cite{LSVQ}, YouTube-UGC \cite{UGC-2019}, and DVL2021 \cite{DVL2021}.
The second type is the IQA dataset, including KonIQ-10k \cite{koniq-10k} and SPAQ \cite{SPAQ}.
The third type is the image classification dataset, including ImageNet-1k and ImageNet-21k \cite{imagenet}.
Their brief information are summarized in Table \ref{dataset}.
It can be observed that these datasets have great difference in content, resolution, time duration, annotation scale, and so on.
Therefore, conducting experiments on these datasets
can reflect the performance of VQA methods from a comprehensive perspective.
Specifically, a subset of LSVQ, named LSVQ-1080p, is constructed. It consists of 3,573 videos, and more than 93\% of them have resolution higher than 1080p.


\subsubsection{Compared methods}
Our proposed StarVQA+ is compared with two types of VQA methods, including non-pretraining methods such as V-BLIINDS \cite{saad2014blind}, VIIDEO \cite{mittal2015completely}, TLVQM \cite{TLVQM} and VIDEVAL \cite{VIDEVAL}, and pretraining methods such as  VSFA \cite{VSFA}, 3D-CNN+LSTM \cite{you2019deep}, MDTVSFA \cite{MDVSFA}, RAPIQUE \cite{tu2021rapique}, PVQ \cite{LSVQ}, LSCT \cite{you2021long}, and BVQA-2021 \cite{li2022blindly}.

\subsubsection{Implementation details}
Our network is built on the PyTorch framework and trained with four Tesla P100 GPUs. During training, unless special indication, the parameters are set as $N=16$, $H=W=224$, $S=16$, $A=12$, and $I=12$.
During testing, a single clip is sampled in the middle of the video.
Three spatial crops (top-left, center, bottom-right) are utilized from the clip and the final prediction is obtained by averaging the scores for these three crops.
The learning rate is initialized as 0.005 and drops by a factor of 0.1 in every 10 epochs with a momentum of 0.9.
All datasets are randomly non-overlapped divided into training set and testing set with a proportion of 8: 2.
For detailed configuration information, please refer to our source code at https://github.com/GZHU-DVL/StarVQAplus.
\begin{figure*}\centering
\setlength{\belowcaptionskip}{-0.18cm}
\subfigure[KoNViD-1k]
{\includegraphics[width=2in]{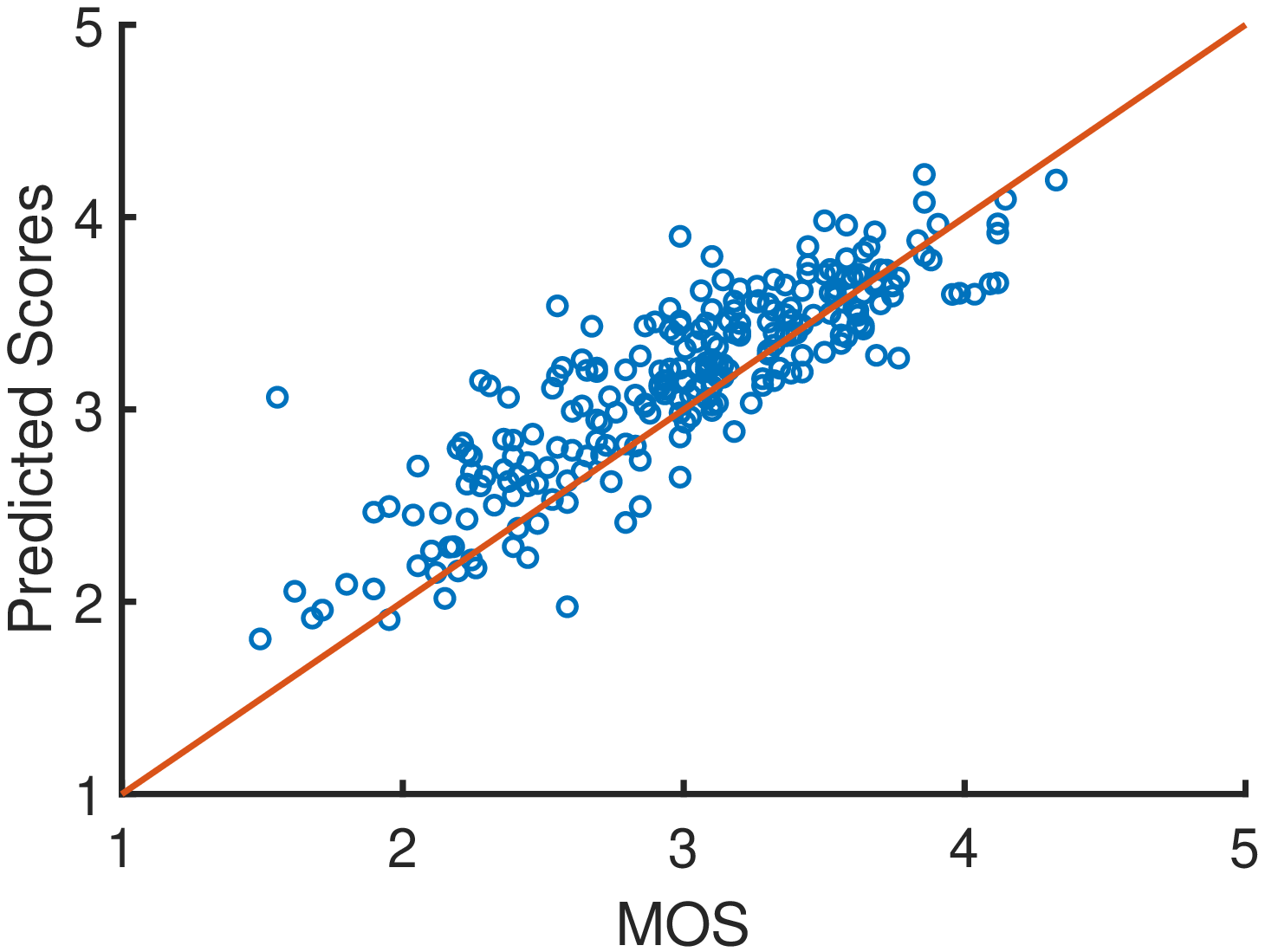}}
\subfigure[LIVE-Qualcomm]
{\includegraphics[width=2in]{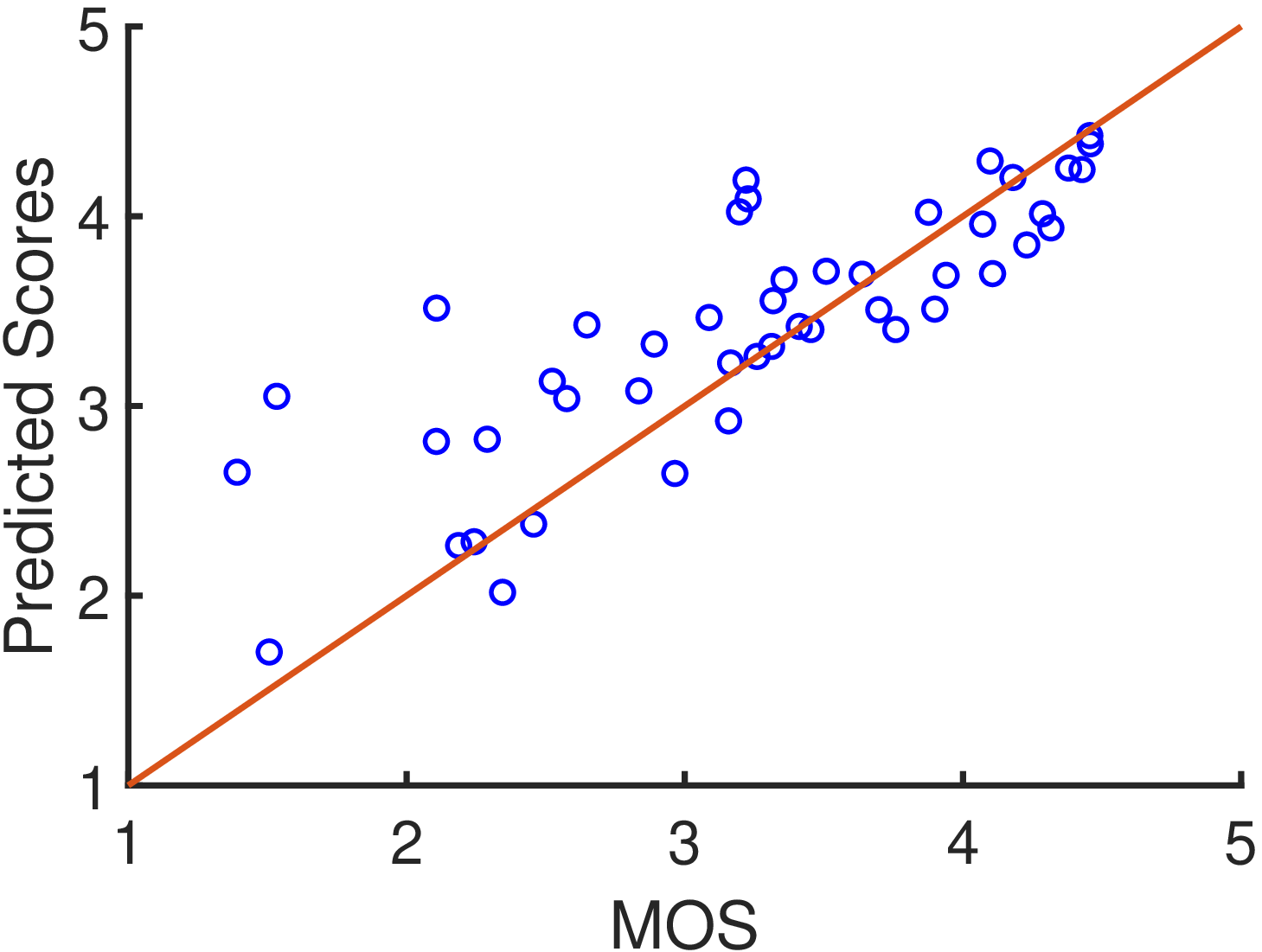}}
\subfigure[LIVE-VQC]
{\includegraphics[width=2in]{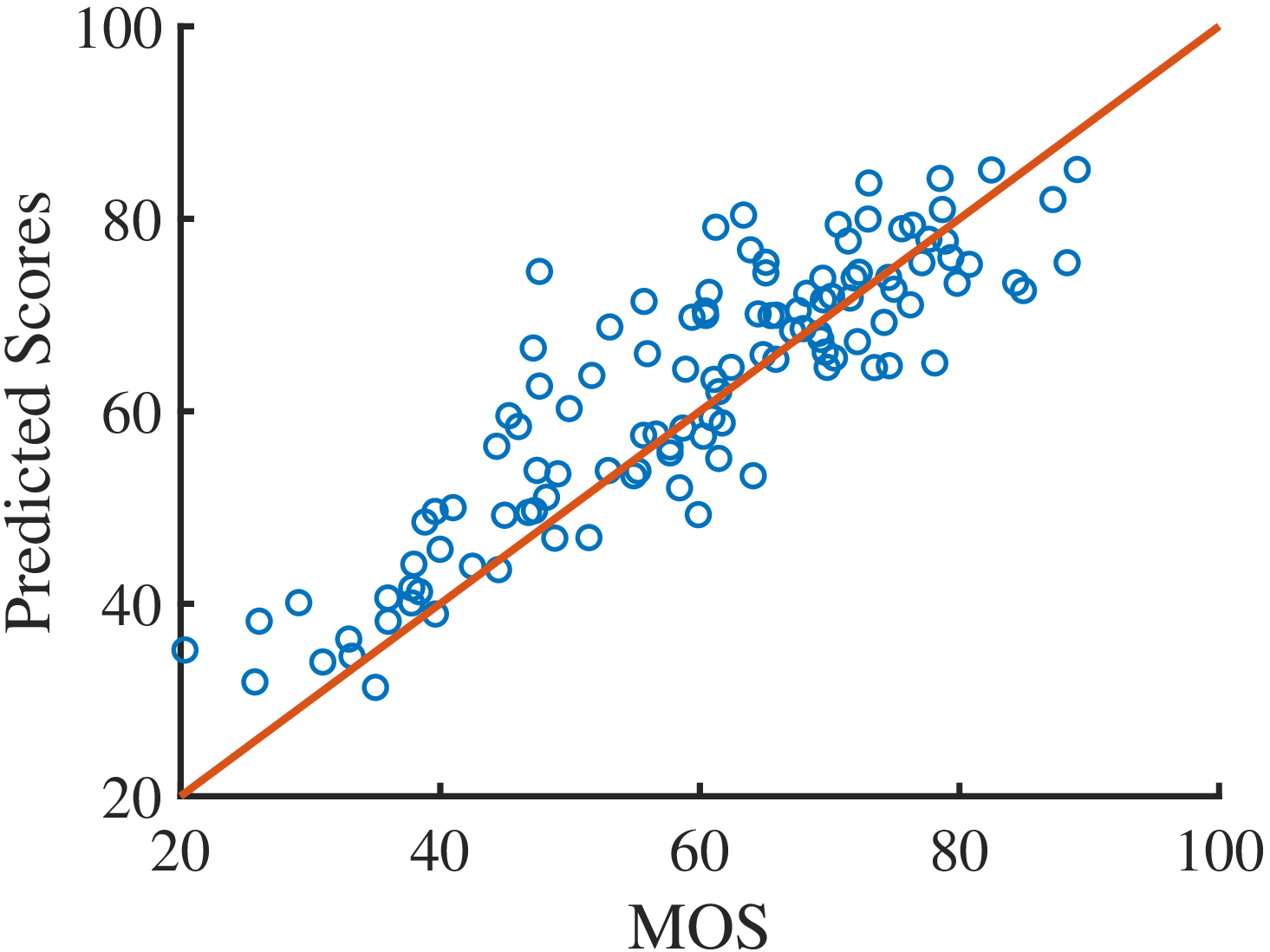}}
\subfigure[YouTube-UGC]
{\includegraphics[width=2in]{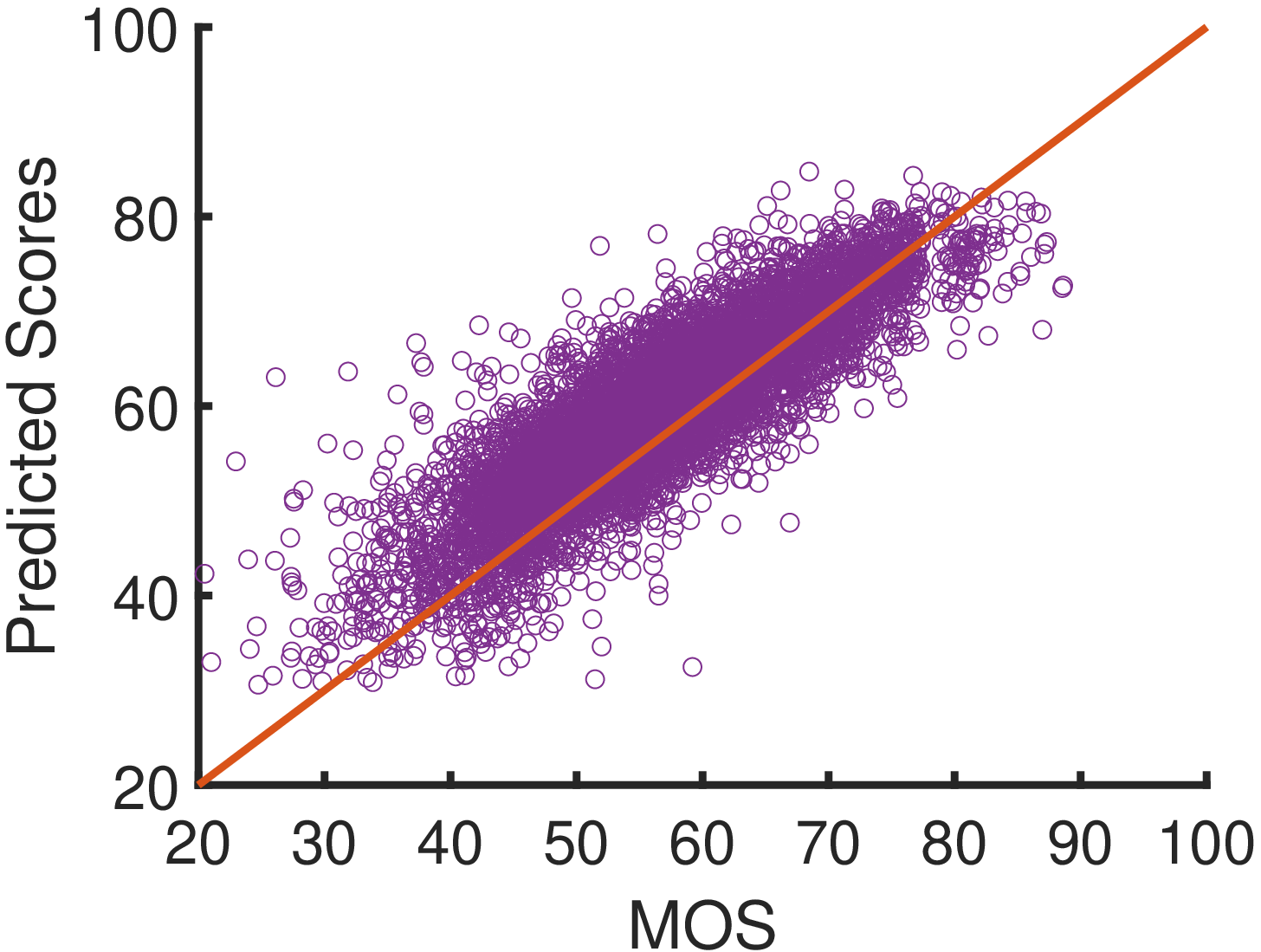}}
\subfigure[LSVQ]
{\includegraphics[width=2in]{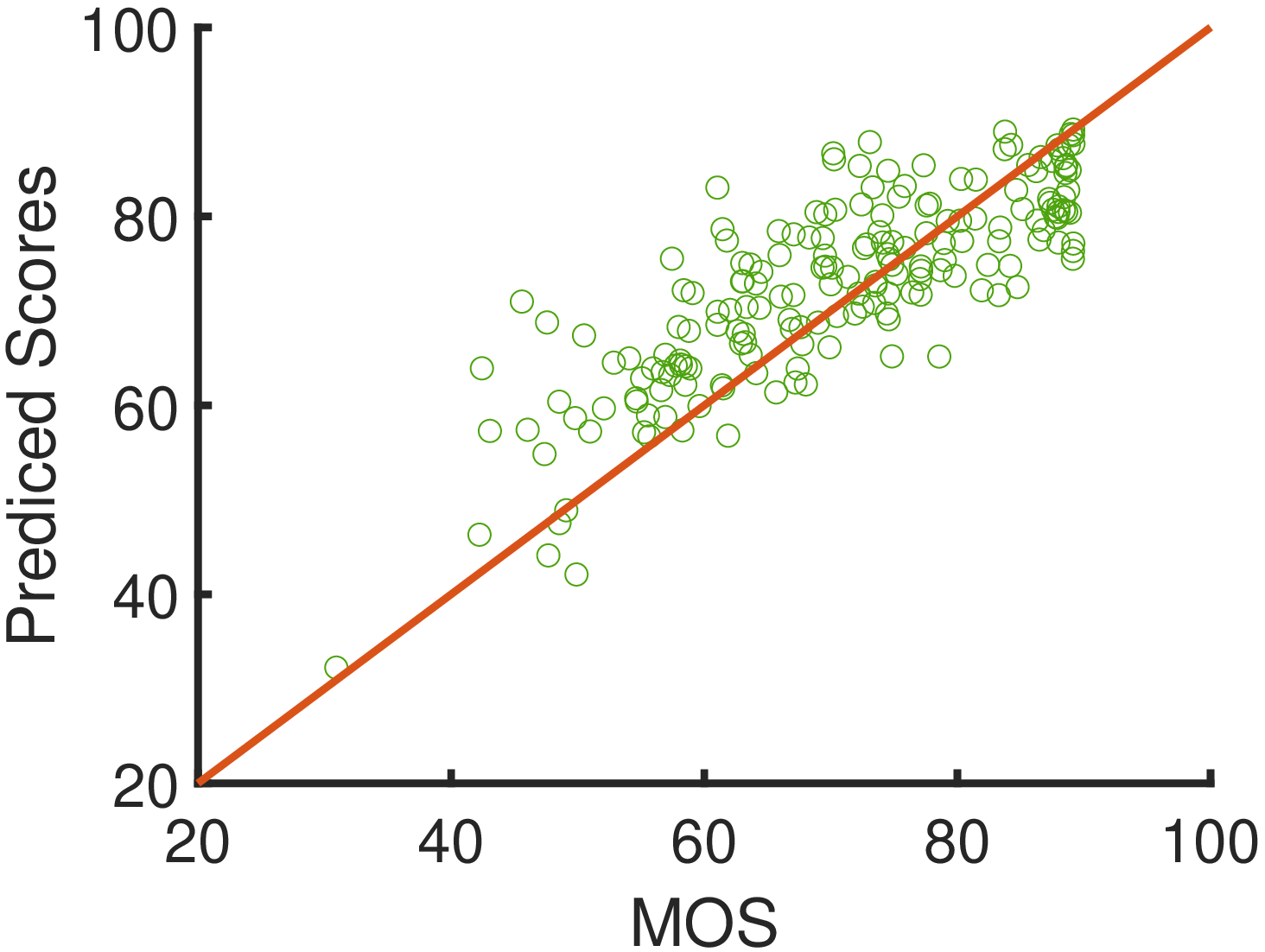}}
\subfigure[DVL2021]
{\includegraphics[width=2in]{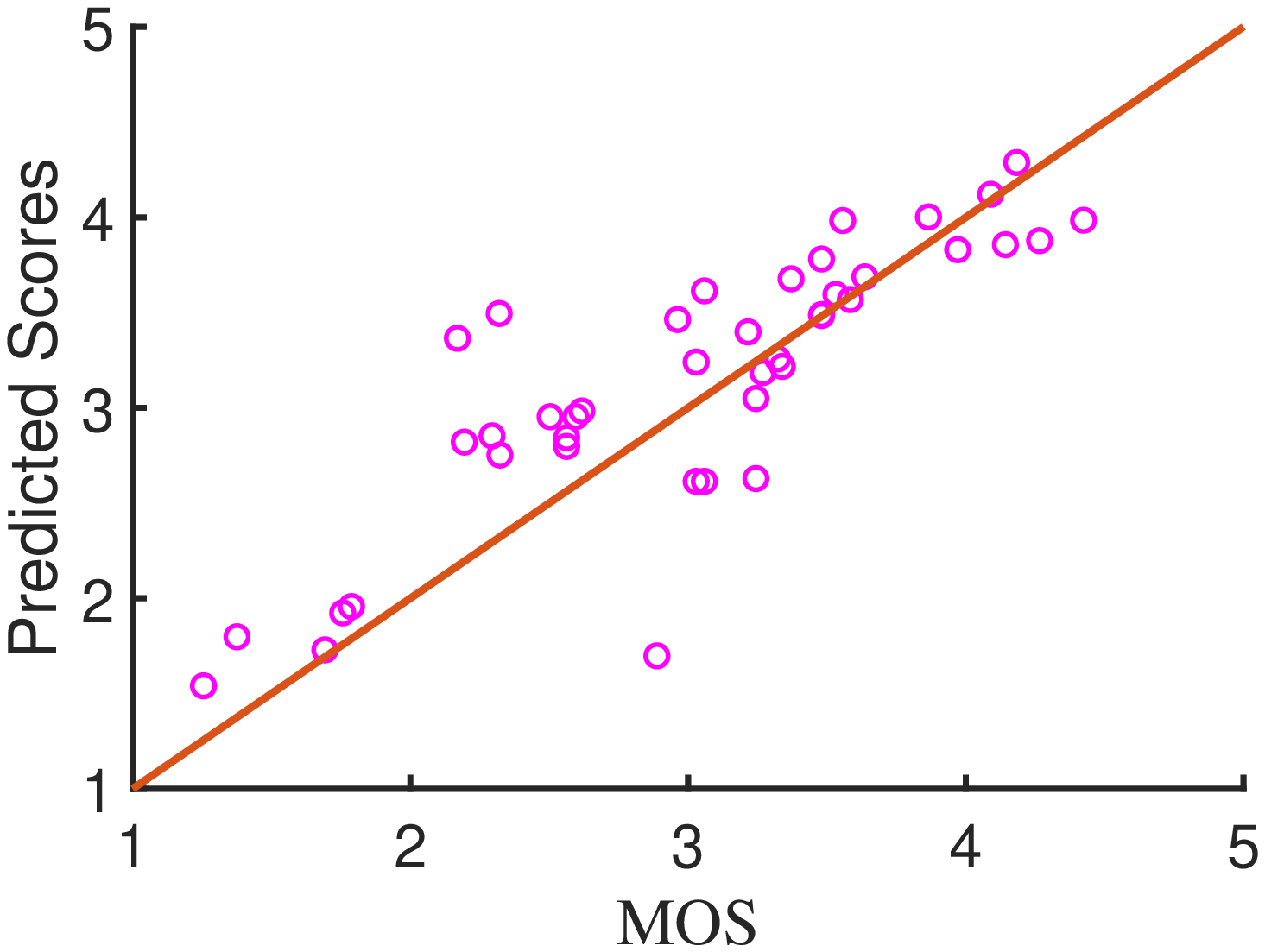}}

\caption{Scatter plots of prediction quality scores output by different dataset. Here, the red line represents the reference line.}\label{scatter}
\end{figure*}

\begin{table*}[htp]

\centering
\caption{Performance comparison on LSVQ} \label{comparison}
\begin{tabular}{lllllll}
\Xhline{1pt} %
 Test Dataset &\multicolumn{2}{c}{LSVQ (7.4K)}&\multicolumn{2}{c}{LSVQ-1080p (3.5k)}&\multicolumn{2}{c}{Weighted average (10.9k)}\\
\hline %
  Models                         &SROCC &PLCC       &SROCC &PLCC        &SROCC &PLCC \\
\hline %
  BRISQUE \cite{BRISQUE}         & 0.5921&0.6382      & 0.6572&0.6581       & 0.5792&0.5761\\

  TLVQM \cite{TLVQM}             &0.7992&0.8031       &0.7731&0.7692        &0.7721  & 0.7741\\

  VIDEVAL\cite{VIDEVAL}          &0.7521&0.7512       &0.7832&0.7802        &0.7941&0.7831\\

  VSFA \cite{VSFA}               &0.7731  & 0.7951    &0.7732 & 0.7753      &0.8012&0.7961\\

  PVQ(w/o) \cite{LSVQ}           &0.8275&0.8376       &0.7915&0.7868        &0.827&0.8286\\

  PVQ(w) \cite{LSVQ}             &0.8271&0.8375       &0.7913&0.7863        &0.8275&0.8281\\

  BVQA-2021 \cite{li2022blindly} &0.8511 &0.8532      &0.7711 &0.7872       &0.8261 &0.8322\\

  StarVQA                        & 0.8512 &0.8571     & 0.8122&0.7926       &0.8391&0.8372\\

  StarVQA+                       & \textbf{0.8532} &\textbf{0.8623}     & \textbf{0.8147}& \textbf{0.8024}       & \textbf{0.8492}&\textbf{0.8421}\\
\Xhline{1pt} %
\end{tabular}
\end{table*}
\begin{table*}[htp]
\caption{The performance evaluation on cross datasets.}\label{Cross-Dataset1}
\centering
\begin{tabular}{l ccc ccc ccc cc}
  \Xhline{1pt} %
{Testing} / {Training}  & \multicolumn{2}{c}{LIVE-Qualcomm (208 samples)} &\multicolumn{2}{c}{LIVE-VQC (585 samples)}   & \multicolumn{2}{c}{KoNViD-1k (1200 samples)}          & \multicolumn{1}{c}{LSVQ (39075 samples)}        \\ \hline
                                  & SROCC    \quad   PLCC  &                & SROCC   \quad PLCC     &          &   SROCC   \quad  PLCC     &        & SROCC   \quad   PLCC                 \\  \cline{2-8}
  LIVE-Qualcomm                   & \textbf{0.6663 \quad 0.6644}&           & 0.6739   \quad 0.6734    &           & 0.6128   \quad 0.6395     &         & 0.5331      \quad 0.6140                  \\
  LIVE-VQC                        & 0.3889 \quad 0.4141&                  & \textbf{0.7316 \quad 0.7303}    & & 0.7805   \quad 0.7752     &         & 0.7801   \quad 0.8054                  \\
  KoNViD-1k                       & 0.3061   \quad 0.3303  &                &0.6724 \quad 0.6842    &         &  \textbf{0.7961   \quad 0.7903}      &          &  0.8420   \quad 0.8577               \\
  LSVQ                           & 0.1995 \quad 0.1902 &                 &0.4722 \quad 0.4798    &         & 0.4550   \quad 0.4960      &          & \textbf{0.8504   \quad 0.8599 }              \\  \Xhline{1pt}

\end{tabular}
\end{table*}
\subsubsection{Performance metrics}
The performance of different methods is evaluated by Spearman's rank ordered correlation coefficient (SROCC) and Pearson linear correlation coefficient (PLCC).
Denote the number of video as $N$, and denote the ground truth quality score and the predicted quality score of the $i$-the video as $y_{i}$ and $\hat{y_{i}}$, respectively.
SROCC is used to measure the monotonic relationship between $y_{i}$ and $\hat{y_{i}}$, and is calculated as
\begin{equation}\label{}
\mathrm{SROCC}=1-\frac{\begin{matrix} 6\sum_{i=1}^N (\hat{y_{i}}-y_{i})^{2} \end{matrix}}{N(N^2-1)}.
\end{equation}
And PLCC is used to measure the linear correlation between $\hat{y_{i}}$ and $y_{i}$, and is calculated as
\begin{equation}\label{}
	\mathrm{PLCC}=\frac{\begin{matrix} \sum_{i=1}^N (y_{i}-\bar{y})(\hat{y_{i}}-\hat{\bar{y}}) \end{matrix}}{\sqrt{\begin{matrix} \sum_{i=1}^N (y_{i}-\bar{y})^{2}\end{matrix}} \sqrt{\begin{matrix} \sum_{i=1}^N (\hat{y_{i}}-\hat{\bar{y}})^{2}\end{matrix}}},
\end{equation}
where $\bar{y}$ and $\hat{\bar{y}}$ denote the mean value of the ground truth and predicted quality scores, respectively.

\subsection{Performance Evaluation on Individual Dataset}

In this part, the performance of our proposed StarVQA+ is evaluated on individual dataset, wherein the training set and the testing set come from the same dataset. The results are shown in Tables \ref{compare-table} and \ref{comparison}, and the following observations can be made

\begin{itemize}
\item[$\bullet$] Our StarVQA+ achieves the state-of-the-art performance on all six different datasets. Compared with the existing best performing method, the average improvement on six datasets is 3.3\% and 2.1\% for SROCC and PLCC, respectively.

\item[$\bullet$] The improvement over the existing best performing method is especially significant on the largest DVL2021 dataset, which is 10.2\% and 2.2\% for SROCC and PLCC, respectively. The reason is that the co-training paradigm in our proposed StarVQA+ can take advantage of the dataset contained a large number of high-definition and ultra-high-definition videos, such as LSVQ and YouTobe-UGC.

\item[$\bullet$] Promisingly, for high-resolution videos, the advantage of the Transformer architecture becomes obvious. From Table \ref{comparison}, it is clearly shown that StarVQA+ surpasses all the competitors when pre-trained on LSVQ and tested on LSVQ-1080p. The reason is that our StarVQA+ may leverage the vast receptive field inside spatial attention to collect spatial information that affects video quality, and can also use the long-term dependency inside temporal attention to extract temporal information that influences video quality.

\item[$\bullet$]
    It can be observed that the performance of the pre-training methods is significantly superior to that of the non-pretraining methods. The reason is that the pre-training stage can tune the network parameters a big step forward.
\end{itemize}


\begin{figure}[tp]
\centering
\includegraphics[scale=0.7]{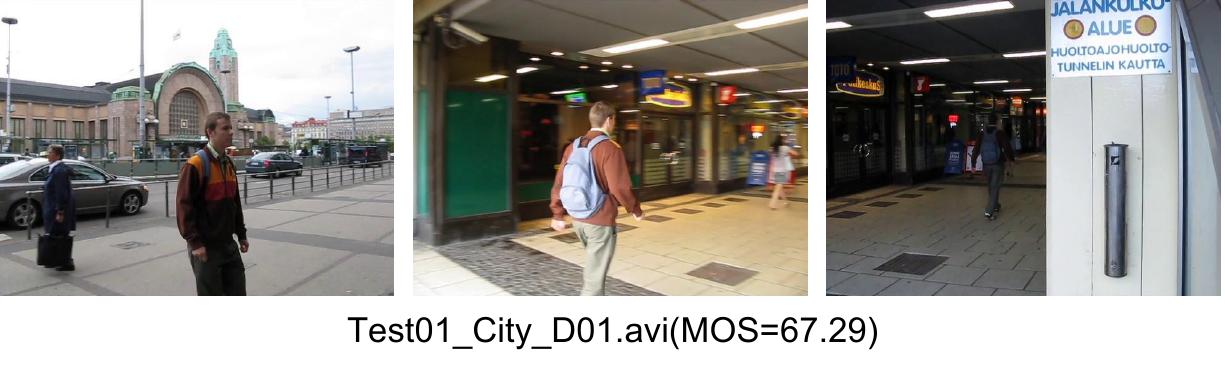}
\includegraphics[scale=0.7]{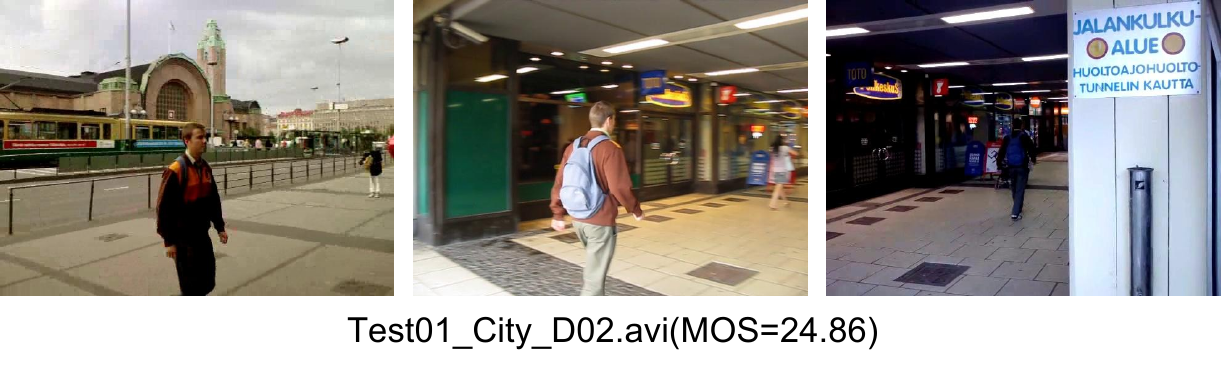}
\includegraphics[scale=0.7]{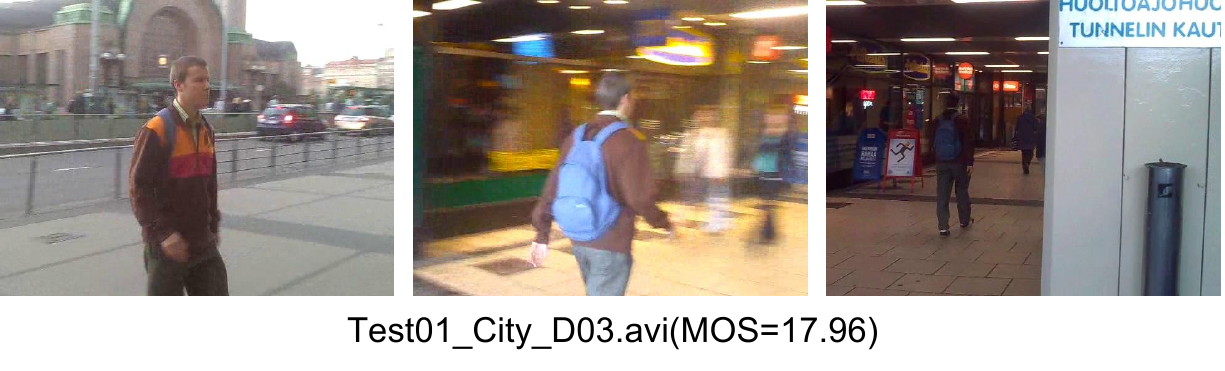}
\includegraphics[scale=0.7]{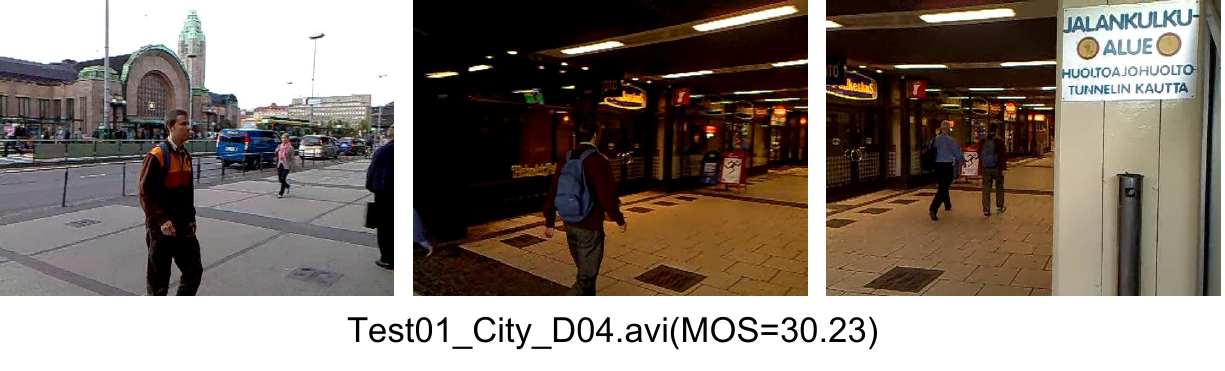}
\includegraphics[scale=0.7]{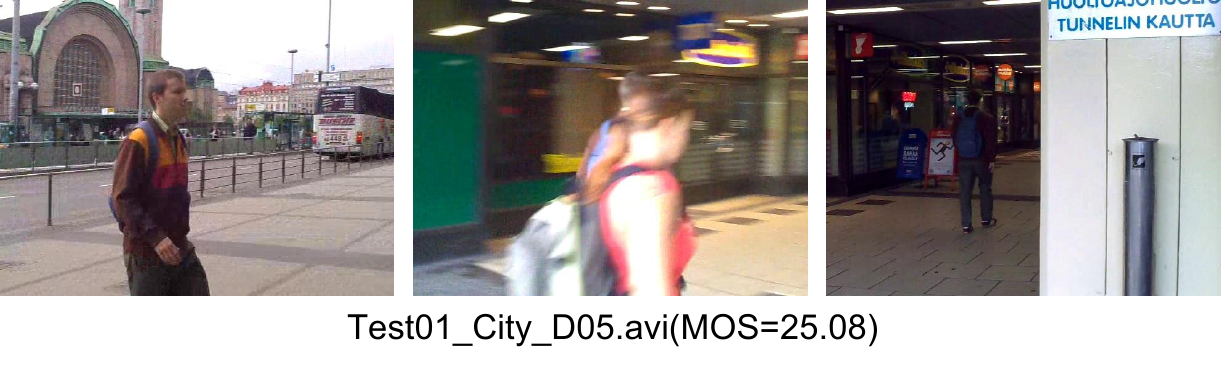}
\includegraphics[scale=0.7]{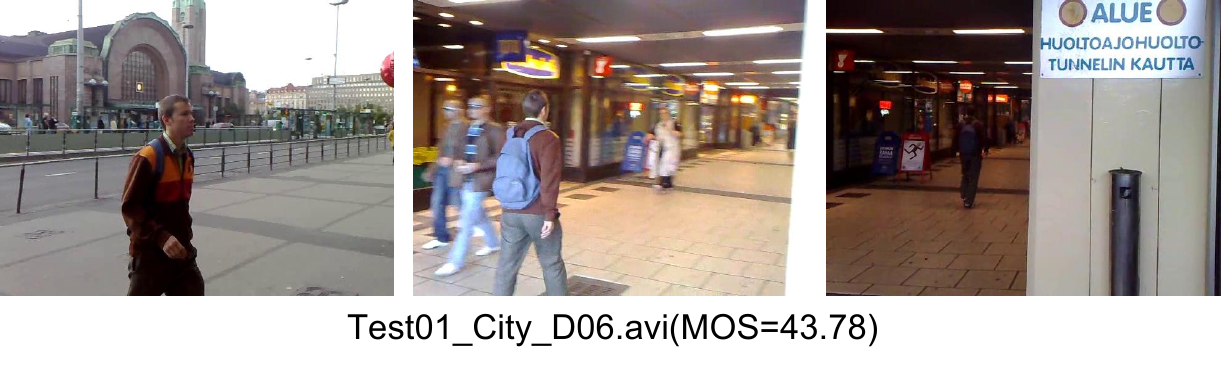}
\includegraphics[scale=0.7]{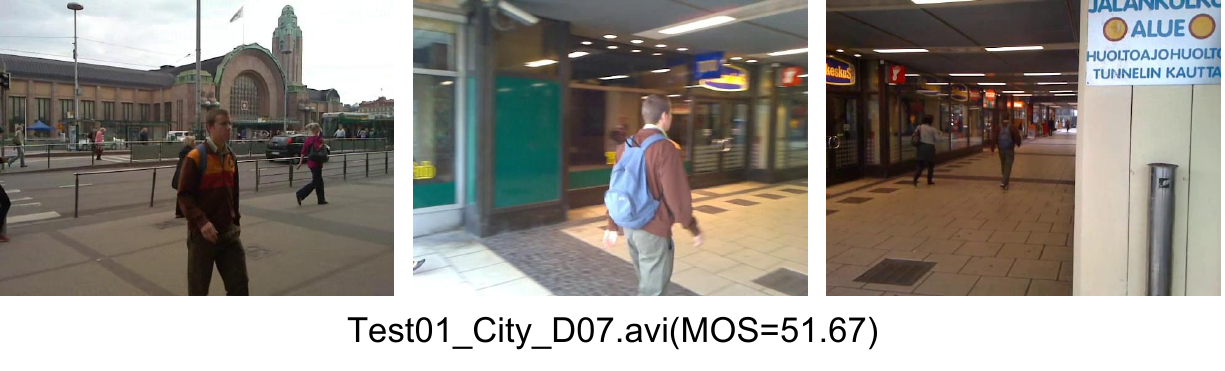}
\includegraphics[scale=0.7]{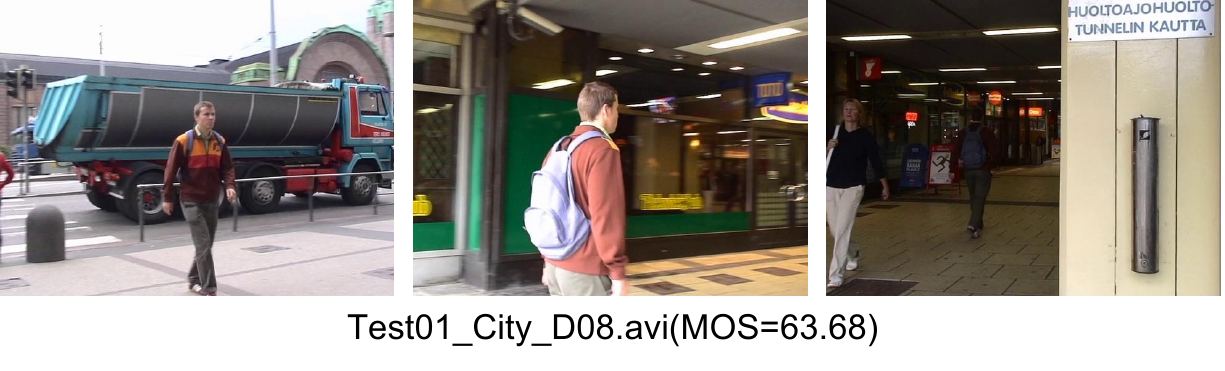}
\includegraphics[scale=0.7]{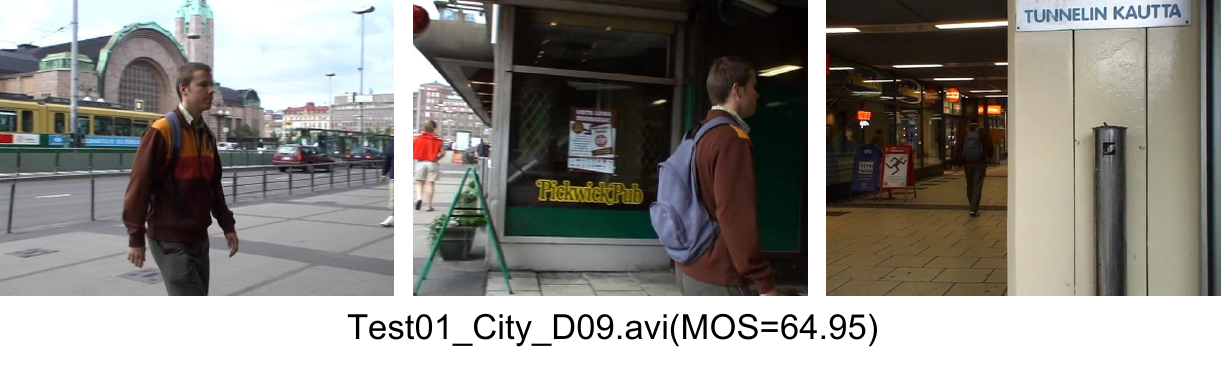}
\caption{The selected frames of the CITY in the TEST01 of the CVD2014 dataset. }\label{CVD2014}
\end{figure}

\subsection{Performance Evaluation on Cross Datasets}
In this part, the performance is evaluated on cross datasets. Since different video datasets have great differences in duration, content, and resolution, evaluations on cross datasets can verify the generalization ability and the robustness of the VQA methods. We conduct two experiments on cross datasets, and the results are given in Table \ref{Cross-Dataset1}, Table \ref{Cross-Dataset2}, and Table \ref{Cross-Dataset3}, respectively. The following observations can be made.

\begin{itemize}
\item[$\bullet$] Table \ref{Cross-Dataset1} shows the results of training models on four datasets. The diagonal position is always the maximum value for this column, and this value becomes larger as the number of training samples increases. This is because the training and testing data are most similar at that position. However, in the case of being trained on LSVQ and tested on LIVE-VQC and KoNViD-1k, we see two anomalous phenomena. The reason is that the LSVQ dataset contains a large number of data samples, rich distortion types, and our network  can have good generalization ability when being fully trained.

\item[$\bullet$] Table \ref{Cross-Dataset2} shows the results of training models on three datasets. It can be observed that our proposed StarVQA+ outperforms existing methods in most cases. Especially, in the case of being trained on LIVE-Qualcomm and tested on LIVE-VQC, KoNViD-1k, LSVQ, StarVQA+ underperforms. The reasons is that our proposed transformer-based architecture has a large number of parameters to train, but LIVE-Qualcomm has too few samples.

\item[$\bullet$] Table \ref{Cross-Dataset3} shows the results of training models on the largest LSVQ dataset. It can be observed that in the case of being tested on KoNViD-1k and LIVE-VQC, the proposed StarVQA+ obtains comparable performance as BVQA-2022.
\end{itemize}

\begin{table*}[htp]
\caption{The performance evaluation on cross datasets.}\label{Cross-Dataset2}
\centering
\begin{tabular}{l ccc| ccc| ccc}
  \Xhline{1pt} %
Training                 & \multicolumn{3}{c|}{KoNViD-1k}                        & \multicolumn{3}{c|}{LIVE-Qualcomm}    & \multicolumn{3}{c}{LIVE-VQC}       \\ \hline
Testing  & LIVE-Qualcomm &  & LIVE-VQC      & KoNViD-1k                        &      & LIVE-VQC         & KoNViD-1k    & & LIVE-Qualcomm         \\   \cline{2-4} \cline{1-7} \cline{8-10}
                          & SROCC \quad   PLCC   &   & SROCC  \quad  PLCC  & SROCC \quad   PLCC    &      & SROCC \quad   PLCC   & SROCC \quad   PLCC && SROCC \quad  PLCC\\ \cline{2-10}
BRISQUE   \cite{BRISQUE}  & 0.3061 \quad 0.3303 &  & 0.5805 \quad 0.5788 & 0.4370 \quad 0.4274  &      & 0.5805 \quad 0.5788 & 0.4370 \quad 0.4274  &      & 0.3601 \quad 0.3303 \\
VSFA     \cite{VSFA}      & 0.5574 \quad 0.5769 &  & 0.6792 \quad 0.7198 & \bf{0.6643 \quad 0.6116}  &      & \bf{0.6425 \quad 0.6819} & 0.6584 \quad 0.6666  &      & 0.5094 \quad 0.5350 \\
TLVQM     \cite{TLVQM}    & 0.4730 \quad 0.5127 &  & 0.5953 \quad 0.6248 & 0.0347 \quad 0.0467  &      & 0.4091 \quad 0.3559 & 0.6023 \quad 0.5943  &      & 0.6415 \quad 0.6534 \\
VIDEVAL   \cite{VIDEVAL}  & 0.4048 \quad 0.4351 &  & 0.5318 \quad 0.5329 & 0.1812 \quad -0.3441 &      & 0.4314 \quad 0.4122 & 0.5007 \quad 0.4841  &      & 0.3021 \quad 0.3602 \\
CNN-TLVQM  \cite{CNN-TLVQM} & 0.6050 \quad 0.6223 &  & 0.7132 \quad 0.7522 & 0.0854 \quad 0.0216  &    & 0.0693 \quad 0.1040 & 0.6431 \quad 0.6304  &      & 0.6574 \quad 0.6696 \\
StarVQA+                   & \textbf{0.6128} \quad  \textbf{0.6395}  & &\textbf{0.7805} \quad  \textbf{0.7752} & 0.3061   \quad 0.3303    &  &   0.3889 \quad  0.4141   & \textbf{0.6724} \quad \textbf{0.6842}  &      & \bf{0.6739 \quad 0.6734}      \\ \Xhline{1pt}
\end{tabular}
\end{table*}

\subsection{Scatter Plot}
To have an intuitive feeling, we visualize the scatter plots between MOS and predicted scores. The scatter plot of predicted quality scores on different datasets is shown in Fig. \ref{scatter}. From first to last, the scatter plots on datasets are KoNViD-1k, LIVE-Qualcomm, LIVE-VQC, YouTube-UGC, LSVQ, and DVL2021, respectively. In each sub-figure, the x-axis indicates the MOS while y-axis indicates predicted score by our method. The scatter points are expected to be located at the diagonal line. The following observations can be made.

 \begin{itemize}

 \item[$\bullet$] The outputs of StarVQA+ are all close to the ground truths. This visually demonstrates that the performance of StarVQA+ remains stable on video sequences from different datasets.
\end{itemize}

 \begin{itemize}

 \item[$\bullet$] Especially, our StarVQA+ on LSVQ gets the most closely centered on the reference line among the three datasets. This is because LSVQ contains a large number of samples, which is most suitable for the Transformer-based networks.
\end{itemize}

We train and test our StarVQA+ on the CVD2014 dataset. However, the loss of our network cannot converge. We analyze the cause as follows. First of all, the CVD2014 dataset has only 234 samples and its samples are divided into 6 tests. There are 3 classes in each test. The video content in each class is very similar, but distortion is quite different. For details, please refer to Fig. \ref{CVD2014}. Our network is pre-trained by ImageNet, thus the extracted features are related to the classification. However, the samples of CVD2014 have similar content but have different MOS, so the loss of our network cannot converge.

\begin{table}[htp]
\centering
\caption{The performance evaluation on cross datasets.}\label{Cross-Dataset3}
\scalebox{1}{
\begin{tabular}{lllll}
\Xhline{1pt} %
Training dataset &\multicolumn{4}{c}{LSVQ}\\
\hline %
Testing datasets &\multicolumn{2}{c}{LIVE-VQC}&\multicolumn{2}{c}{KoNViD-1k}\\
\hline %
Models &SROCC &PLCC &SROCC &PLCC \\
\hline

  BRISQUE \cite{BRISQUE} & 0.524&0.536& 0.646&0.647\\

  TLVQM \cite{TLVQM}&0.670  & 0.691 &0.732  & 0.724 \\

  VIDEVAL \cite{VIDEVAL} &0.630&0.640&0.751&0.741 \\

  VSFA \cite{VSFA} &0.734&0.772&0.784&0.794\\

  PVQ \cite{LSVQ} &0.770 &0.807&0.791&0.795\\

  BVQA-2022 \cite{li2022blindly} &\bf{0.816} &\bf{0.824}&0.839&0.830 \\
  StarVQA+  &0.7853&0.806&\bf{0.842}&\bf{0.858}             \\
 \Xhline{1pt} 
\end{tabular}

}
\end{table}
\begin{table}[htp]
\centering
\caption{Performance under different loss functions.} \label{loss function}
\begin{tabular}{lll}
\Xhline{1pt}
Test Dataset& \multicolumn{2}{c}{KoNViD-1k} \\
\hline
 Loss               &SROCC & PLCC  \\
\hline
$L_2$ Loss          & 0.771  & 0.820 \\

Cross Entropy Loss  & 0.811 &0.811  \\

Our Loss            & \bf{0.863}  & \bf{0.862}  \\
\Xhline{1pt}

\end{tabular}
\end{table}
\begin{table}[htp]
\centering
\caption{Performance comparison under different numbers of frames. The top performance is highlighted in boldface.}\label{frame_number}
\begin{tabular}{lclll}
\Xhline{1pt}
Training Dataset &\multicolumn{3}{c}{KoNViD-1k}\\
\hline
Pretrain Dataset & Frame Number & SROCC     & PLCC  \\
\midrule
\multirow{5}{*}{Dataset Combination-6}
    & 2    &0.8210   & 0.8345 \\

    & 4    &0.8354    & 0.8480\\

    & 8    &0.8520   & 0.8613\\

    & 16   &\bf{0.8763}  & \bf{0.8809}\\
    & 24    & 0.8594     & 0.8669  \\
    & 32    &0.8683      & 0.8747 \\
\Xhline{1pt}
\end{tabular}
 \end{table}

\begin{table}[htp]
\centering
\caption{Performance comparison under different anchor points. The top performance is highlighted in boldface.}\label{anchor number}
\begin{tabular}{lclll}
\Xhline{1pt}
Training Dataset &\multicolumn{3}{c}{KoNViD-1k}\\
\hline
Space-Attention & Anchor Number & SROCC & PLCC  \\
\hline
\multirow{5}{*}{Dataset Combination-6}
   & 3   &0.8334    &0.8211\\

   & 6  &\bf{0.8763}    &\bf{0.8809} \\

   & 11  &0.8745    &0.8775 \\

   & 21   &0.8740    &0.8740\\

   & 51  &0.8728      &0.8683\\
\Xhline{1pt}
\end{tabular}
\end{table}

\begin{table*}[htp]
\centering
\caption{The complexity comparison of our method with several mainstream VQA methods. }\label{efficiency}
\begin{tabular}{l|lc|cc|cc}
\Xhline{1pt}
\multirow{2}{*}{Methods} &\multicolumn{2}{c|}{540p}    &\multicolumn{2}{c|}{720p}  &\multicolumn{2}{c}{1080p}  \\ \cline{2-7}
                    & FLOPs (G)      &Time (s)                 &FLOPs (G)       &Time (s)                  &FLOPs (G)       &Time (s) \\  \hline
 VSFA               & 10249         & 2.603                  & 18184         &3.571                    & 40919         &11.14   \\
 PVQ                & 14646         & 3.091                  & 22029         &4.143                    & 58501         &13.79  \\
 BVQA-2022          & 28176         & 5.392                  & 50184         &10.83                    & 112537        &27.64   \\
 StarVQA+           & 590           &0.212                   &590            &0.483                    & 590           &0.884         \\
\Xhline{1pt}
\end{tabular}
\end{table*}
\subsection{Performance Comparison with Different Loss}
To better illustrate the role of our designed loss function, we conduct an experiment to compare different loss functions. The experimental results are shown in Table \ref{loss function}. We can see that the VR loss function gets the best performance. The reason is that our VR loss combines the characteristics of the classification problem and the regression problem, making full use of classification networks to solve regression problems.

\subsection{Hyperparameter Study} \label{anchor}
To verify the rationality of each hyperparameter of the proposed method, in this part, we conduct a series of studies from the following two aspects.

First, we study the number of frames sampled from each video sample, and we take different values for $N$, including 2, 4, 8, 16, 24, 32. Then, a comparative experiment is carried out under the same other conditions and different frame numbers. Second, for the VRloss proposed by us, we have also carried out experimental verification on the discussion of the number of anchors in VRloss. We took 5 groups of anchor points for verification. The number of anchor points is 3, 6, 11, 21, 51. The experimental results are shown in Table \ref{frame_number} and Table \ref{anchor number}. The following observations can be made.

\begin{itemize}
\item[$\bullet$] From the experimental results in Table \ref{frame_number}, we can see that with the continuous increase of the number of frames, the obtained performance is also continuously improved. When $F$ takes 16, the performance reaches the maximum. When we increase the frame number again, we found that there is a small drop in performance. From the above results, we conclude that when the duration of the video sequence is 8 seconds, we only need to sample 16 frames uniformly from this sequence and our StarVQA+ can get good performance. Compared with many methods that require frame-by-frame training \cite{VSFA, you2019deep, MDVSFA, tu2021rapique, you2021long, li2022blindly}, our method can achieve huge advantage in saving the training resource.

\item[$\bullet$] It can be seen from Table \ref{anchor number} that when the number of anchor points is 3, the obtained performance is the worst. When the number of anchor points is 6, the performance is the best. If we continue to increase the number of anchors, we can see that the performance basically remains unchanged. So throughout the experiment, the number of our anchor points is set to 6.
\end{itemize}

\subsection{Computational Complexity of StarVQA+}
In this part, we compare the FLOPs and running times on GPU (average on ten runs per sample) of the proposed method with existing deep VQA approaches on different resolutions to demonstrate the computational efficiency of StarVQA+. The experimental results are shown in Table \ref{efficiency}. We can observe that for various resolutions, our method achieves the lowest computational complexity in terms of FLOPs and run time. All these comparisons show the great computational efficiency of the proposed StarVQA+.

\section{Conclusion}
This paper has developed a novel space-time attention network for VQA, named StarVQA+. To our best knowledge, we are the first to apply pure-Transformer to the VQA task. Furthermore, a new vectorized regression loss function has designed to adapt the Transformer-based architecture for training. To overcome the limitation of insufficient labeled data, a co-training paradigm has been proposed to train the huge volume of Transformer network parameters. Experimental results show that the proposed StarVQA+ achieves competitive performance compared with mainstream VQA methods. This work broadens Transformer to a new application and demonstrates that the self-attention has excellent potential in the VQA field.

\bibliographystyle{ieeetr} \small 
\bibliography{Refs}

\end{document}